\documentclass{article} % For LaTeX2e
\usepackage[submission,final]{colm2026_conference}

\usepackage{microtype}
\usepackage{hyperref}
\usepackage{url}
\usepackage{booktabs}
\usepackage{lineno}

\usepackage{graphicx}
\usepackage{algorithm}
\usepackage{algorithmic}
\usepackage{amsmath}
\usepackage{amssymb}
\usepackage{mathtools}
\usepackage{amsthm}
\usepackage[capitalize,noabbrev]{cleveref}
\usepackage{wrapfig}

\newcommand{\jb}[1]{ {\color{blue}[JB: #1]}}

\newcommand{\action}[1]{\texttt{\##1}}

\usepackage{etoc}
\usepackage{dsfont}
\usepackage{siunitx}
\usepackage[table]{xcolor} 
\usepackage{pifont}
\usepackage{multirow}
\usepackage{rotating}

\usepackage{epigraph} 
\usepackage{enumitem}
\usepackage{comment}
\usepackage[font=small]{caption}

\usepackage{tikz}
\usepackage{pgfplots}
\usepackage{tikzscale} % <--- The magic package
\usepackage{standalone} % Ignores the preamble inside the included .tex files
\usepgfplotslibrary{groupplots}
\usetikzlibrary{calc}
\pgfplotsset{compat=1.18}

\usepackage[most,skins,theorems]{tcolorbox}

\tcbset{
  aibox/.style={
    width=\linewidth,
    top=6pt,
    bottom=0pt,
    colback=blue!6!white,
    colframe=black,
    colbacktitle=black,
    enhanced,
    center,
    attach boxed title to top left={yshift=-0.1in,xshift=0.15in},
    boxed title style={boxrule=0pt,colframe=white},
  }
}
\newtcolorbox{AIbox}[2][]{aibox,title=#2,#1}

\definecolor{safeRed}{RGB}{213, 94, 0}      % Vermilion
\definecolor{safeBlue}{RGB}{0, 114, 178}    % Strong Blue
\definecolor{safeGreen}{RGB}{0, 158, 115}   % Bluish Green

\pgfplotscreateplotcyclelist{ocean}{
    {fill=safeRed, draw=none},
    {fill=safeBlue, draw=none},
    {fill=safeGreen, draw=none},
}

\tcbuselibrary{listings, breakable}
\tcbset{
    graybox/.style={
        colback=gray!20,  % Background color (light gray)
        colframe=black,   % Border color (black)
        boxrule=1pt,      % Border thickness
        arc=4mm,          % Box corner rounding
        width=\textwidth, % Box width to match the text width
        before=\vspace{10pt}, % Space before the box
        after=\vspace{10pt},  % Space after the box
        center             % Center the box
    }
}

\definecolor{chatblue}{RGB}{0, 0, 255}
\definecolor{chatred}{RGB}{255, 0, 0}
\definecolor{darkblue}{rgb}{0, 0, 0.5}

\def\ambigqa{\textsc{AmbigQA}}
\def\pacific{\textsc{Pacific}}

\def\ourapproach{\textsc{SGP}}
\def\ouroracle{\textsc{SGP-Oracle}}

\definecolor{darkblue}{rgb}{0, 0, 0.5}
\hypersetup{colorlinks=true, citecolor=darkblue, linkcolor=darkblue, urlcolor=darkblue}

\title{Learning Steerable Clarification Policies with Collaborative Self-play}

\author{
  \textbf{Jonathan Berant$^{1\dagger}$\thanks{Corresponding author: \texttt{joberant@google.com}}, Maximillian Chen$^{1\dagger}$, Adam Fisch$^{2\dagger}$, Reza Aghajani$^3$,} \\
  \textbf{Fantine Huot$^2$, Mirella Lapata$^2$ and Jacob Eisenstein$^{2\dagger}$} \\
  \normalfont{$^1$Google Research, $^2$Google DeepMind, $^3$Meta, work done at Google Deepmind} \\
  \normalfont{$^\dagger$Core contributor}
}

\begin{document}

\ifcolmsubmission
\linenumbers
\fi

\maketitle

\begin{abstract}
To handle ambiguous queries, AI assistants must decide (a) when to guess the user's intent and answer directly, (b) when to enumerate and respond to multiple plausible intents, and (c) when to ask a clarifying question. 
No strategy is always superior; the right decision depends on \emph{contextual} factors such as user preferences or modality. For example, enumerating several possible interpretations can be cumbersome on small screens or voice-based settings, while a more interactive style may be dispreferred in desktop settings. In this work, we propose training  \emph{steerable policies} for managing such uncertainty via self-play. 
Given two agents, one simulating a user and the other an AI assistant, we generate conversations where the user issues a potentially ambiguous query, and the assistant must choose how to respond. The assistant receives as input the \emph{numerical cost} of asking a clarification question, and of generating each word, and is trained to select the conversational action that maximizes its final reward:  accuracy penalized by these costs. We apply Reinforced Self-Training (ReST) to optimize this objective and show that it yields a steerable policy that adapts its behavior based on the provided costs, improving both reward and accuracy. Moreover, the resulting policy generalizes to numerical cost values unobserved during training.

\end{abstract}

\section{Introduction}
\label{sec:intro}

Establishing common ground between a user and an AI assistant is key for successful communication, but is an area where AI assistants have been shown to struggle~\citep{shaikh-etal-2024-grounding}.
As a result, assistants may offer responses that fail to match the true intent of an underspecified user request, or offer overly verbose responses that attempt to accommodate all possible intents --- both failure modes that could be avoided by more interactive policies for building common ground, e.g., through clarification questions. 
Recent work \citep{chen2025learningclarify,zhang2025modelingfutureconversationturns} has proposed to teach models when to ask clarification questions by using \emph{self-play}: an episodic learning strategy in which one language model simulates a user, and another represents the AI assistant. The assistant is then finetuned on successful ``rolled-out'' conversations to learn when to answer a query and when to ask for clarification.

However, such prior work assumes that ambiguity should be uniformly handled by requesting clarification, which does not capture the richness of strategies a conversational participant might take to manage uncertainty. Given an ambiguous utterance, there are many possible approaches: (a) guess a single probable intent and respond to it; (b) enumerate multiple possible intents and provide a response for each; (c) issue a clarification question to pinpoint the exact intent. 
Which strategy is optimal? This depends on whom you are talking to, and in what context. Some users dislike long outputs and favor a more conversational approach; others prefer to avoid multi-turn conversations, but will relax this preference when it is not possible to process long outputs, such as while driving. In low-stakes settings it may be acceptable to simply make a best guess. A proficient assistant should therefore take preferences, interaction modality, and other environmental factors into account when deciding how to respond. Thus, we need \emph{steerable policies for conversation grounding}, which let the assistant flexibly adjust its dialogue strategy on the fly.

%Assuming users want to get a correct answer as quickly as possible, determining the optimal policy requires weighing the cost of each strategy: 
%the cost of more conversation turns for clarification questions, the cost of generating long responses when accommodating multiple interpretations, and the probability of getting the wrong answer when guessing. Importantly,  these costs are \emph{context-dependent}, varying based on  user preference, interaction modality, and scenario. For example, when using speech while driving, a long response enumerating multiple query interpretations can waste valuable time and  lead to cognitive overload. Conversely, in a laptop it is easier to skim different interpretations without bothering to answer a clarification question. The importance of developing \emph{steerable policies for conversation grounding} has been by and large absent from prior work on handling user ambiguity. 

\begin{figure}[!t]
    \centering
    \includegraphics[width=0.63\linewidth]{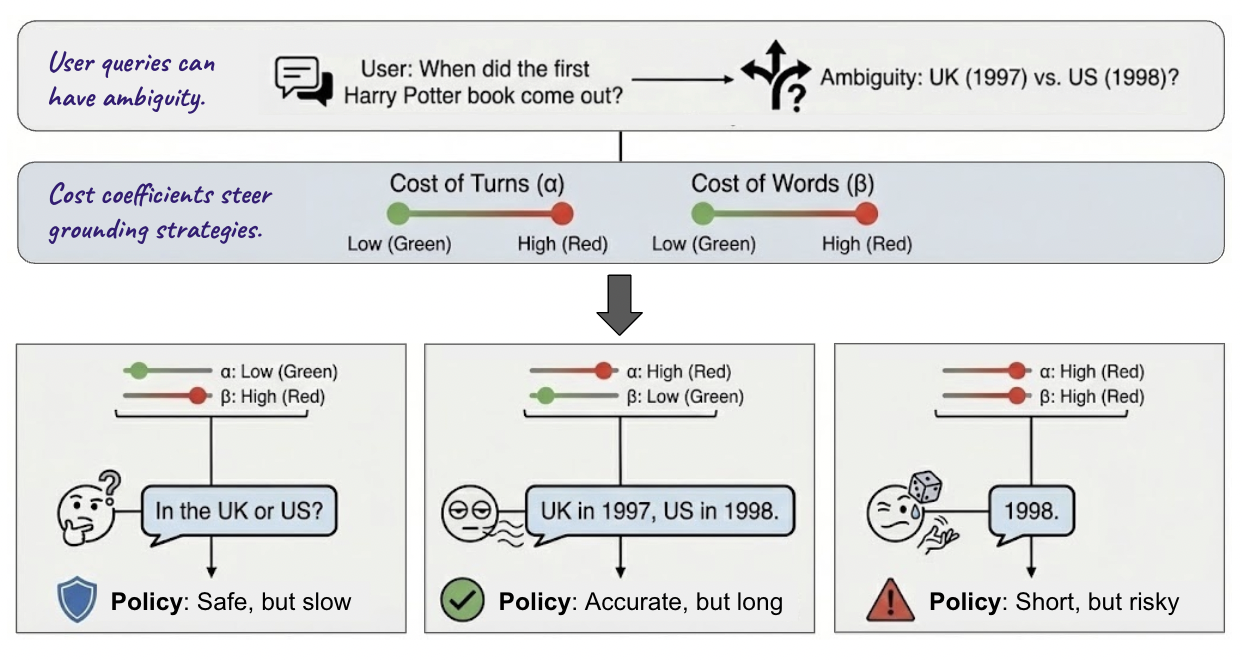}
    \caption{Our steerable AI assistant  adapts its behavior based on cost coefficients for different dimensions of the conversation. The model is trained to maximize \emph{cost-penalized accuracy}, with costs specified by $\alpha$ and $\beta$ (the color red indicates high cost and the color green indicates low cost).
    In this example, the model asks a clarification question when the cost of conversation turns is low (\emph{left}), enumerates multiple interpretations when the cost of generating a long response is low (\emph{middle}), and answers directly with an educated guess when both costs are high (\emph{right}).}
    \label{fig:intro}
    \vspace{-10pt}
\end{figure}

We propose a new form of self-play for training a single \emph{steerable} policy that takes as input
a specification of the reward for a conversation, and modifies its grounding behavior to maximize the provided reward. \Cref{fig:intro} provides an example for an ambiguous query. 
The model is trained to maximize \emph{cost-penalized accuracy}, where cost depends on the number of turns and the length of the final answer. This objective is parameterized by \emph{scalar cost coefficients}, $\alpha$ and $\beta$, provided as part of the input. % The model adapts its behavior based on these coefficients. 
On the left, the cost of additional conversation turns is low, so the model clarifies uncertainty by asking a follow-up question. In the middle, the cost of enumerating interpretations is low, so the model provides answers for two probable interpretations. On the right, both costs are high, so the model makes an educated guess about user intent, without resolving the uncertainty. We focus in this work on training the assistant for \emph{steerability}, assuming cost coefficients are given as input. In practice, cost coefficients can be inferred from user preference data or  feedback, which we leave for future work.\looseness=-1

To finetune our model, we apply Reinforced Self-Training (ReST), a common post-training strategy for optimizing rewards~\citep[e.g.,][]{gulcehre2023reinforced,zhang2024rest}. We use one model as a user simulator and another as our target assistant, sampling different cost coefficients per conversation. We then apply collaborative self-play. For each question and cost coefficient setting, we sample multiple conversations, and compute the true reward. We then finetune the assistant on the highest-reward conversations, teaching it to condition on the cost coefficients to maximize its reward.

We evaluate our Steerable Grounding Policy (\ourapproach{}) on two question-answering benchmarks: \ambigqa{}~\citep{min-etal-2020-ambigqa}, where the model answers questions using parametric knowledge, and \pacific{}~\citep{deng-etal-2022-pacific}, where non-parametric context is provided in the form of a document and table.
We find that even strong prompted models fail to condition on cost coefficients, even when using chain-of-thought prompting. In contrast, models \emph{trained} with \ourapproach{} appropriately condition on the provided cost coefficients; when needed, they can substantially alter their inference-time behavior, achieving higher accuracy at lower cost in accordance with the objective specified in the prompt. Moreover, this generalizes to cost coefficients never observed during training.

\vspace{-10pt}
\section{Method}
\label{sec:method}

% We start by describing the self-play environment and inference mechanism. Then, we present our training procedure.
\vspace{-5pt}
\subsection{Self-play Environment}
\label{subsec:inference}

Our self-play setup includes two agents, a \emph{user simulator} (or \emph{user} for short) and an \emph{assistant}, who are collaborating to answer a \emph{query} provided by the \emph{environment}. The environment has access to a
potentially ambiguous query $q$, a list of possible interpretations $\mathcal{I}_q = (i_1, \dots, i_n)$, a corresponding distribution over the interpretations $\mathcal{P}_q=(p_1, \dots, p_n)$, where $\forall_j p_j \geq 0, \sum_j p_j=1$, and the gold answers $(a_1, \dots, a_n)$. The environment samples an interpretation~$i_j$, where $j \sim \mathcal{P}_q$ and passes it along with $q$ to the user simulator.
The user simulator is instructed to determine the answer by communicating with an \emph{assistant} (who may have access to privileged information). After the ensuing multi-turn conversation, the user returns an answer $a$ to the environment, which is compared with the 
gold answer $a^* = a_j$. An example conversation is shown in \Cref{fig:rollout}.

% \begin{figure}[!t]
%     \centering
%     \includegraphics[width=1.\linewidth]{figures/example_rollout_nano.png}
%     \caption{An example rollout. The environment samples an interpretation for an ambiguous query~$q$ and passes~$q$ and its unambiguous interpretation~$i$ to the user. The user passes~$q$ to the assistant, which issues a clarification question, and after receiving a clarification response, outputs a multi-answer (answer that covers multiple interpretations), which is used to formulate the final answer.}
%     \label{fig:rollout}
%     \vspace{-10pt}
% \end{figure}

\begin{wrapfigure}{r}{0.5\textwidth} % {r} for right side, {l} for left
    \centering
    \vspace{-10pt} % Pulls the figure up to align better with text
    \includegraphics[width=0.48\textwidth]{figures/example_rollout_nano.png}
    \caption{An example rollout. The environment passes the user an ambiguous query $q$ and unambiguous interpretation $i$. The user passes~$q$ to the assistant, which issues a clarification question. After receiving a response, it outputs a multi-answer (covering multiple interpretations), from which the user formulates the final answer.}
    \label{fig:rollout}
    \vspace{-15pt} % Reduces space below the caption
\end{wrapfigure}

\noindent
\textbf{Rollouts.} A conversation, or \emph{rollout}, is a sequence of turns $\rho = ((M_1, x_1, y_1), \dots, (M_T, x_T, y_T))$, where each turn includes the active model (user or assistant) $M_t$, the prompt $x_t$ constructed from information in the state (see Table~\ref{tab:agent_actions}), and the model output, $y_t$. For us, $M_t$ is the user when $t$ is odd, and the assistant when $t$ is even.

The output $y_t = (u_t, o_t)$ includes a chosen \emph{action} and an observation for that action. The prompt $x_1$ is constructed based on the query $q$ and the sampled interpretation $i_j$. The rollout terminates when the user returns the final answer $a = o_T$ alongside $u_T = \action{FINALIZE}$.

\noindent
\textbf{Agent actions and observations.}
Table~\ref{tab:agent_actions} defines all actions for the user and assistant, along with the state, used to construct the prompt $x_t$, and the output observation.
The assistant can choose one of three actions: \{\action{CLARIFY}, \action{ANSWER}, \action{MULTI\_ANS}\}, corresponding to asking a clarification question, answering the query, or providing multiple interpretations. Our  goal is to train the assistant to choose actions that maximize its reward (see below). The user chooses actions in a deterministic manner: in the first turn, it will \action{QUERY};
when given a clarification question $q_\textrm{clar}$ it will \action{RESPOND} with a clarification response $r_\textrm{clar}$; and when given an answer or multi-answer from the assistant it will generate  and return a final answer $a$ (\action{FINALIZE}). When the assistant gives a multi-answer, the user
leverages knowledge of $i_j$ to choose the relevant part of the multi-answer.
In practice we limit the number of allowed clarification questions (see~\Cref{sec:experiments}). %An example rollout is shown in Figure~\ref{fig:rollout}.

\noindent
\textbf{Reward.} Let $n_\textrm{clar}(\rho)$ denote  the number of clarification actions and $|o_{T-1}|$  the number of words in the final assistant answer. The rollout reward computed by the environment is
{\setlength{\abovedisplayskip}{3pt}
\setlength{\belowdisplayskip}{3pt}
\begin{equation*}
    R(\rho; a^*) = \textrm{acc}(a, a^*) - \alpha \cdot n_\textrm{clar}(\rho)- \beta \cdot |o_{T-1}|
\end{equation*}
We want the answer to be accurate, to minimize the number of turns, and to get short answers. Our reward is motivated by Grice's maxims~\citep{grice1975logic} for effective communication, including the maxim of \emph{quality} (make your contributions true),  \emph{quantity} (provide necessary information only), and \emph{manner} (avoid ambiguity).
We measure accuracy, $\textrm{acc}(a, a^*)$, against the gold answer $a^*$ with token-level F$_1$~\citep{rajpurkar-etal-2016-squad}, in a $[0,100]$ scale.

The optimal strategy for this reward function depends on whether it is possible to increase expected accuracy with clarification questions or multi-answers. E.g., if asking a clarification question can increase expected accuracy by~$\delta$, and $\alpha < \delta$, then asking will yield higher reward compared to answering directly. Similarly, if a multi-answer adds $m$~words to the final assistant answer but improves expected accuracy by $\delta$, it is preferable when~$\beta \cdot m < \delta$.

\noindent
\textbf{User and assistant prompts.}
Prompts $x$ are constructed from the state, as explained in Table~\ref{tab:agent_actions}. The assistant gets the coefficients $\alpha$ and $ \beta$ so it can learn to change its behavior conditioned on them.
The assistant has access to the history of clarification questions and their answers $h = \{(q_\textrm{clar}^1, r^1_\textrm{clar}), \dots\}$, and to an optional \emph{private context} $c$. This is used when the query is w.r.t to some context (e.g., a table) that can be provided in the prompt. We show an example where the assistant has access to a private context in~ \Cref{fig:nonparametric_knowledge} (Appendix~\ref{app:pacific_examples})
% \jb{MAX, please add this} 
and describe the details in Section~\ref{sec:experiments}.
The full prompts are provided in Appendix~\ref{app:prompts}.

We rely on the base model's instruction-following ability in two ways: (a)~given a clarification question, we instruct the user to answer without revealing the true interpretation and (b)~given the answer~$r_\textrm{ans}$, we instruct the user simulator to only use $r_\textrm{ans}$ and its knowledge of the true interpretation $i_j$ to generate the final answer~$a$. The user is instructed to never answer based on its parametric knowledge.
We manually verify that the user's instruction-following behavior is satisfactory. 
In addition, using parametric knowledge will lead to lower performance in asymmetric cases where the assistant's knowledge is superior, e.g., when the assistant has access to a private context containing necessary knowledge.

\noindent
\textbf{Thoughts.}
We experiment with models that use chain-of-thought reasoning to generate actions. Before sampling an action, the model generates a thought that considers all allowed actions, and estimates which will yield the highest reward (similar in spirit to~\cite{dong-etal-2026-value}).
Our hypothesis is that using a chain-of-thought will help the model better condition on the input cost coefficients, since it will explicitly reference coefficients to estimate the reward for each action.
We provide the model a few in-context examples with thoughts, shown in the full prompts in \Cref{app:ai_assistant_prompt}.

\subsection{Training}
\label{subsec:training}

\begin{wraptable}{r}{0.53\textwidth}
\centering
\vspace{-15pt} % Reduces space below the caption
\small
\begin{tabular}{@{}ll|lc@{}}
\toprule
\textbf{Agent} & \textbf{Action} & \textbf{State}  & \textbf{Observation} \\
\midrule
User & \action{QUERY} & $q, i$  & $q$ \\
 & \action{RESPOND} & $q, i, q_\textrm{clar}$  & $r_\textrm{clar}$ \\
 & \action{FINALIZE} & $q, i, r_\textrm{ans}$  & $a$ \\
 \midrule
Assistant & \action{CLARIFY} & $q, h, c, \alpha, \beta$  & $q_\textrm{clar} $ \\
& \action{ANSWER} & $q, h, c, \alpha, \beta$ &  $r_\textrm{ans} $ \\
& \action{MULTI\_ANS} & $q, h, c, \alpha, \beta$ &  $r_\textrm{ans} $ \\
\bottomrule
\end{tabular}
%\medskip
\caption{
The user observes the query $q$, its interpretation $i$, and potentially (a) a clarification question $q_\textrm{clar}$, (b) answer $r_\textrm{ans}$ or (c) multi-answer $r_\textrm{ans}$.
The assistant observes the query, a history $h$ of clarification questions and answers, an optional private context $c$, and the coefficients $\alpha, \beta$.
The assistant outputs (a) clarification questions $q_\textrm{clar}$ or (b) answers or multi-answers $r_\textrm{ans}$.
The user responds to clarification questions with $r_\text{clar}$, and synthesizes a final answer $a$ given $r_\textrm{ans}$.
 The state is used to construct the prompt $x$. \looseness=-1
}
\vspace{-15pt}
\label{tab:agent_actions}
\end{wraptable}

Our goal is to train a steerable assistant that maximizes reward given input cost coefficients. To this end, we apply Reinforced Self-Training (ReST), a post-training method that approximates on-policy learning by updating model parameters between each epoch of training~\citep[e.g.,][]{gulcehre2023reinforced,zhang2024rest}. In ReST, 
one generates rollouts, scores them, trains on high-scoring rollouts, and then iterates with the updated policy.
However, the highest-scoring rollout might feature poorly-chosen actions that obtained a high reward by chance, e.g. by guessing an intent that is underspecified but (luckily) correct. To avoid training on such noisy supervision, we seek the \emph{action sequence} that leads to the highest reward in expectation over rollouts and interpretations, and construct training examples from the best rollout compatible with this action sequence (see also \citealt{eisenstein2025dontliefriendslearning,zhang-choi-2025-clarify}). \Cref{alg:rest} describes the
implementation.

Recall that a rollout is defined as $\rho = \{(M_t, x_t, y_t)\}_{t=1}^T$, with action-observation outputs $y_t = (u_t, o_t)$. An action sequence $\mathbf{s} = (s_1, s_2, \ldots)$ is a sequence of actions (e.g., \action{ANSWER} or \action{CLARIFY}). If for a given rollout $\rho$ we have $u_t = s_t$ for all $t$, we write $\rho \vdash \mathbf{s}$ to indicate that $\rho$ is \emph{compatible} with $\mathbf{s}$. A set of rollouts, $\{ \rho_1, \rho_2, \ldots \}$ can be coarsened into a set of action sequences $\{ \mathbf{s}_1, \mathbf{s}_2, \ldots \}$. Given a query $q$ and coefficients $\alpha, \beta$, we can look at
their set of rollouts, coarsen the rollouts to action sequences and identify the action sequence, $\mathbf{s}_*^q$ with the highest expected reward across interpretations (we drop $\alpha$ and $\beta$ for brevity):
\begin{align}
\mathbf{s}_*^q &= \underset{\mathbf{s}^q}{\operatorname{argmax}} \overbrace{\sum_{j \in [n]} p_j \cdot \underbrace{\sum_{\{\rho: \rho \vdash \mathbf{s}^q\}} p_{t-1}(\rho|q,i_j) \cdot R(\rho)}_{\mathbb{E}_\rho[R(\mathbf{s}^q \mid q, i_j)]}}^{\mathbb{E}_\rho[R(\mathbf{s}^q \mid q)]} 
\end{align}
where $p_{t-1}$ corresponds to the probability of the trajectory using the current assistant $M^\mathcal{A}_{t-1}$ and user $M^\textrm{user}$. In practice, we assume $\mathcal{P}_q$ is uniform and estimate $p_{t-1}$ through sampling: we compute average reward per interpretation and average across interpretations, choosing the action sequence that does well across all of them.
Once we have $\mathbf{s}_*^q$, 
we go over all interpretations, and add the highest-reward rollout compatible with  $\mathbf{s}_*^q$ to the training set:
\begin{align}
\rho_*^q &= \bigcup_{i_j \in \mathcal{I}_q}\underset{\{\rho : \rho \vdash \mathbf{s}_*^q, i(\rho) = i_j\}}{\operatorname{argmax}} R(\rho),
\end{align}
where $i(\rho)$ is the interpretation used in the trajectory $\rho$.

\begin{algorithm}[tb]
\caption{Reinforced Self-Training}

\label{alg:rest}
\begin{algorithmic}
\scriptsize % Using small instead of scriptsize for better readability
\REQUIRE Initial assistant model $M^\mathcal{A}_0$, fixed user model $M^\textrm{user}$, query set $\mathcal{Q} = \{(q_k, \mathcal{I}_{q_k}, \mathcal{P}_{q_k})\}_{k=1}^N$, set of coefficients $\mathcal{C} = \{(\alpha_j, \beta_j)\}_{j=1}^M$, rollouts per query $n_r$, number of ReST iterations $T$.
\ENSURE Trained model $\mathcal{M}^{\mathcal{A}}_{\textrm{ReST}}$

\FOR{$t = 1, \dots, T$}
    \STATE Define an empty training set $\mathcal{D} \leftarrow \emptyset$.
    \FOR{query triple $(q_k, \mathcal{I}_{q_k}, \mathcal{P}_{q_k}) \in \mathcal{Q}$}
        \STATE Sample index $j \sim \mathcal{P}_{q_k}$ and interpretation $i_j \in \mathcal{I}_{q_k}$.
        \FOR{$\ell = 1, \dots, n_r$}
            \STATE Sample $\alpha, \beta \sim \mathcal{C}$.
            \STATE Generate and save rollout $\rho_{k,j, \ell}$ by applying models $M^\mathcal{A}_{t-1}$ and $M^\textrm{user}$ to $q_k, i_j, \alpha, \beta$.
        \ENDFOR
    \ENDFOR
    \FOR{query $q_k$, its rollouts $\rho_k$, and $(\alpha, \beta) \in \mathcal{C}$, }
        \STATE \textbf{Step a:} Filter out invalid rollouts (see Section~\ref{subsec:training}).
        \STATE \textbf{Step b:} Compute best action sequence:
        \STATE \quad $\mathbf{s}_*^{k, \alpha,\beta} = \arg\max_{\mathbf{s}^{k, \alpha, \beta}} \mathbb{E}_\rho[R(\mathbf{s}^{k, \alpha, \beta} \mid q_k)]$
        \STATE \textbf{Step c:} Compute best rollouts compatible with $\mathbf{s}_*^{k, \alpha,\beta}$:
        \STATE \quad $\rho_*^{k, \alpha, \beta} = \bigcup_{i_j \in \mathcal{I}_{q_k}} \arg\max_{\{\rho : \rho \vdash \mathbf{s}_*^{k, \alpha, \beta}, i(\rho) = i_j\}} R(\rho)$
        \STATE \textbf{Step d:} Turn $\rho_*^{k, \alpha, \beta}$ into training examples, add to $\mathcal{D}$.
        %turn-level training examples and add to $\mathcal{D}$.
    \ENDFOR
    \STATE Train $\mathcal{M}^{\mathcal{A}}_{t}$ on a training split of $\mathcal{D}$.
    \STATE Select checkpoint with smallest held-out log-likelihood on a development split of $\mathcal{D}$.
\ENDFOR
\STATE \textbf{return} $\mathcal{M}^\mathcal{A}_T$
\end{algorithmic}
\end{algorithm}

% Specifically, for each query $q$, interpretation of the query $i_j$, and cost coefficients $\alpha,\beta$, we identify \emph{sequences of actions} that reliably lead to good rewards. Then, we select the best \emph{rollout} compatible with such action sequences (see Algorithm~\ref{alg:rest}). Recall that a rollout is defined as $\rho = \{(M_t, x_t, y_t)\}_{t=1}^T$, with action-observation outputs $y_t = (u_t, o_t)$. An action sequence $\mathbf{s} = (s_1, s_2, \ldots)$ is a sequence of actions (such as \action{ANSWER} or \action{CLARIFY}). If for a given rollout $\rho$ we have $u_t = s_t$ for all $t$, then we write $\rho \vdash \mathbf{s}$ to indicate that $\rho$ is \emph{compatible} with $s$. Now a set of rollouts, $\{ \rho_1, \rho_2, \ldots \}$ can be coarsened into a set of action sequences $\{ \mathbf{s}_1, \mathbf{s}_2, \ldots \}$. Each action sequence is scored by the average reward of its compatible rollouts, and the best action sequence $\mathbf{s}_*^{\alpha,\beta}$ is selected for each setting of cost coefficients, reducing the noise in sampling individual rollouts. We then choose the best individual rollout $\rho_*^{\alpha,\beta} \vdash \mathbf{s}^*$, and train on the model assistant steps of this rollout.

\Cref{alg:rest} describes this procedure.
We generate~$n_r$ rollouts for each query~$q_k$ and  interpretation, sampling cost coefficients in each rollout.
We cluster the $n_r$ rollouts by cost coefficients, filtering out invalid rollouts. A rollout is invalid if its F$_1$ is below some threshold, or if the user simulator violates instructions and ignores the assistant's answer. This is determined by observing that the fraction of words in $o_T$ that also appear in $o_{T-1}$ is below a threshold. Finally, for each query $q_k$ and coefficient pair $(\alpha,\beta)$ we 
find the optimal action sequence $\mathbf{s}_{*}^{k, \alpha,\beta}$ 
and add to the training set 
the best rollouts $\rho_{*}^{k, \alpha,\beta}$ that are compatible with $\mathbf{s}_{*}^{k, \alpha,\beta}$, across interpretations of $q_k$.
We only train on the \emph{assistant} steps. 

Crucially, different settings of $\alpha$ and $\beta$ lead to different optimal action sequences. This lets the model learn the association between coefficient values and desired behaviors. 
% For example, rollouts with clarification questions are less likely to maximize reward when $\alpha$ is high.
ReST training proceeds for~$T$ epochs, where we sample from the latest model in each epoch.

\vspace{-10pt}
\section{Experimental Setup}
\label{sec:experiments}

% \subsection{Datasets}
% \label{sec:datasets}

% Colab for processing AmbigQA
% https://colab.corp.google.com/drive/1b1F-hR8iUDJZq2Mco9DMnBU9y6CR3R-Q#scrollTo=H8_zasNXgrBH
%We evaluate on two benchmarks:~\ambigqa~\cite{min-etal-2020-ambigqa} and \pacific~\cite{deng-etal-2022-pacific}. 

\paragraph{Datasets} \ambigqa{} is a QA benchmark built from Natural Questions~\citep{kwiatkowski-etal-2019-natural}, where queries are annotated with multiple interpretations and their answers. In manual inspection, we found noisy annotations (annotated interpretations that are not plausible). We therefore perform model-based filtering: we ask Gemini 2.0 Flash~\citep{gemini2} to determine if a question is ambiguous, and keep only queries where models agree with human annotations (see \Cref{app:filtering_prompt} for our prompt). Then, we subsample the unambiguous queries such that the ratio of ambiguous to unambiguous queries is 2:1. Overall, this leads to 1,776 training examples and 382 development (dev) examples (the test set is not publicly available), with an average of~2.41 interpretations per query on the training set and 2.76 on the dev set. An example ambiguous query is \emph{``How many basketball national championships does Louisville have?''}, referring to men's or women's basketball.

\pacific~\citep{deng-etal-2022-pacific} is a multi-turn QA task that requires reasoning about contextual ambiguity grounded in real financial documents and tabular data (see \Cref{fig:nonparametric_knowledge}).
This creates two differences from \ambigqa{}. First, this is an \emph{asymmetric} setup, where the assistant provides an answer that can only be derived from an input document, which the user has no access to. This  prevents the user from answering questions using its own knowledge while ignoring the assistant's response.
% \jb{max - can you stitch together what I wrote with what you wrte? we probably need an example figure here or in an appendix.}
Second, the context-grounded and multi-turn nature of \pacific{} introduce different types of ambiguity. For example, a turn that appears underspecified may not be ambiguous given previous context (e.g., red text in \Cref{fig:multiturn-ambiguity}). Conversely, seemingly precise queries can be ambiguous due to details in the  context (e.g., blue text in \Cref{fig:multiturn-ambiguity}). We use 3,744 training examples and 640 dev examples (test set ground truth is not publicly available), with equal numbers of ambiguous and unambiguous queries.

% \subsection{Experimental Setup}
% \label{sec:baselines}

\paragraph{Models} 
For the main results presented here,  we use Gemma 2 9B~\citep{gemmateam2024gemma2improvingopen} as our baseline model for both the user simulator\footnote{We tested sensitivity to the user simulator, by using Gemma 3 4B as the user simulator instead of Gemma 2 9B, and found results were similar.}
and AI assistant. 
In Appendices~\ref{app:gemma_2_2b} and~
\ref{app:qwen_3_4b}, we show additional results with Gemma 2 2B and QWEN 3 4B.
We compare several natural baselines to our model:

\begin{itemize}[leftmargin=*, noitemsep, nolistsep, topsep=0pt]
\item \textsc{Prompted}: queries the baseline model directly with a description of the task, reward objective, and several examples. This tests whether the model can natively condition on cost coefficients. We show the prompt in \Cref{app:ai_assistant_prompt}.
\item \textsc{Prompted-COT}: Similar to \textsc{Prompted}, except that the prompt includes instructions and examples to generate a thought that explicitly reasons about which action will maximize reward. %(see \Cref{sec:method}). 
See the prompt in \Cref{app:ai_assistant_prompt}.
\item \textsc{Answer}:  uses the baseline but constrains it to  answer directly without clarification questions or multi-answers.
\item \textsc{Multi-answer (MA)}: uses the baseline but constrains the assistant to use \action{MULTI\_ANS}.
\item \textsc{Clarify}: uses the baseline but constrains the assistant to use \action{CLARIFY} and then \action{ANSWER}.
\item \textsc{Clarify-and-MA}: uses the baseline but constrains the assistant to use \action{CLARIFY} and then \action{MULTI\_ANS}.
\item \textsc{Answer-FT}: To teach the base model the right answer format, we sample direct answers per question and finetune on the highest-reward answer observations.
\item \ourapproach{}: trains a Steerable Grounding Policy (\Cref{alg:rest}).
%\item \jb{Additional possible baselines: (a) always ask clarification questions (b) Finetuned answer only for learning to output good strings. (c) Finetune a different model for each $alpha$ $beta$ value as an upper bound (d) more?}
\end{itemize}

We also include \ouroracle{}, an oracle that provides an approximate upper bound on the achievable reward. Using \ourapproach{}, for every query $q$ and coefficients $\alpha, \beta$, we estimate the expected reward $\mathbb{E}_\rho[R(\mathbf{s} \mid \alpha, \beta)]$ for each action sequence $\mathbf{s}$. We then select the sequence, $\mathbf{s}^*$, that in hindsight attains the highest expected reward. This offers a reference point for the potential gains from optimal action sequences. 
%achievable by choosing the best possible action sequence.

Last, we conduct a few ablations related to the details of our training procedure, which we describe in Appendix~\ref{app:ablations}.

\paragraph{Training and inference details}
We use $n_r=192$ rollouts at training time and $n_r=64$ at test time.
Interpretations are sampled uniformly from the candidates provided in \ambigqa{} and \pacific{}. In each rollout we randomly sample the clarification cost coefficient $\alpha \sim \{0, 2, 20\}$ and the per-token cost coefficient $\beta \sim \{0.1, 0.7, 5.0\}$. We limit the number of clarification questions to $1$, since typically there is just one ambiguous aspect in each query, which results in four possible action sequences: answer vs. multi-answer; with and without clarification questions.
We perform $T=3$ ReST epochs, following prior work~\citep{eisenstein2025dontliefriendslearning}. We tune only the learning rate, choosing the checkpoint with lowest dev loss.
%where we average the NLL for actions and observations.
As explained in \Cref{subsec:training}, we discard rollouts with low F$_1$: $<0.1$ for \ambigqa{} and $<0.4$ for \pacific{}. We also discard rollouts when the user ignores the assistant, omitting rollouts when the fraction of tokens in the answer returned by the user that do not appear in the assistant answer is above half. We provide statistics on how often filtering happens in Appendix~\ref{app:filter_stats}.

\paragraph{Evaluation}
We use the following  metrics: (a) average reward, (b) average F$_1$,
(c) fraction of queries with a clarification question, (d) fraction of queries that use \action{MULTI\_ANS}. We compare models by query type (ambiguous and unambiguous) and across cost coefficients. 
We also report metrics for \emph{steerability}, which we describe in detail below. Last, we measure steerability generalization by testing coefficients that did not appear during training.

\begin{table}[t]
\centering
\scriptsize
\begin{tabular}{@{}llr|>{\columncolor{gray!20}}c>{\columncolor{gray!20}}r>{\columncolor{gray!20}}r|r|>{\columncolor{gray!20}}c>{\columncolor{gray!20}}r>{\columncolor{gray!20}}r@{}}
% \toprule
& \multirow{2}{*}{\textbf{Models}} & \multicolumn{4}{c}{\textbf{Ambiguous}} & \multicolumn{4}{c}{\textbf{Unambiguous}} \\
\cmidrule(lr){3-6} \cmidrule(lr){7-10}
& & \textbf{Reward} & \textbf{F1} & \textbf{\% Clar.} & \textbf{ \% MA} & \textbf{Reward} & \textbf{F1} & \textbf{\% Clar.} & \textbf{\% MA} \\
\cmidrule(lr){2-10}
%\multicolumn{9}{@{}l}{\textbf{AmbigQA}} \\
%\midrule
&\textsc{Prompted} & -34.49 & 16.16 & 66.83 & 63.88 & -15.01 & 33.07 & 64.78 & 58.07 \\
&\textsc{Prompted-COT} & -20.11 & 16.44 & 53.93 & 50.48 & -5.02 & 29.12 & 48.34 & 49.54 \\
\cmidrule(lr){2-10}
&\textsc{Answer} & 1.32 & 14.33 & 0.00 & 0.00 & 26.73 & 35.58 & 0.00 & 0.00 \\
&\textsc{Clarify} & -9.59 & 16.30 & 100.00 & 0.00 & 8.86 & 31.86 & 100.00 & 0.00 \\
&\textsc{Multi-answer (MA)} & -46.48 & 15.67 & 0.00 & 100.00 & -25.40 & 33.76 & 0.00 & 100.00 \\
&\textsc{Clarify-and-MA} & -50.61 & 16.94 & 100.00 & 100.00 & -35.26 & 32.28 & 100.00 & 100.00 \\
&\textsc{Answer-FT} & 3.64 & 16.40 & 8.75 & 9.98 & 30.99 & 39.65 & 2.31 & 4.58 \\
\cmidrule(lr){2-10}
&\ourapproach{} & \textbf{6.63} & \textbf{19.94} & 40.25 & 18.22 & \textbf{31.36} & \textbf{41.28} & 26.46 & 13.99 \\
\cmidrule(lr){2-10}
\raisebox{1cm}[0pt]{\begin{sideways}{\textbf{AmbigQA}}\end{sideways}}&\ouroracle{} & 15.81 & 23.13 & 35.78 & 11.08 & 42.05 & 48.66 & 20.03 & 10.32 \\ \midrule
\midrule
% \multicolumn{10}{@{}l}{} \\ \midrule
&\textsc{Prompted} & 11.46 & 55.28 & 17.99 & 47.18 & 42.44 & 69.51 & 4.82 & 19.15 \\
&\textsc{Prompted-COT} & 33.00 & 50.04 & 16.45 & 16.17 & 52.07 & 68.43 & 4.86 & 7.17 \\
% &\textsc{Prompted-Gemma 3 4B UserSim} & 6.81 & 53.62 & 18.64 & 42.58 & 37.15 & 70.00 & 4.17 & 17.86 \\
\cmidrule(lr){2-10}
&\textsc{Answer} & 42.44 & 52.25 & 0.00 & 0.00 & 56.91 & 71.52 & 0.00 & 0.00 \\
&\textsc{Clarify} & 45.48 & 62.70 & 100.00 & 0.00 & 46.90 & 68.85 & 100.00 & 0.00 \\
&\textsc{Multi-answer (MA)} & -21.39 & 54.63 & 0.00 & 100.00 & -4.04 & 63.34 & 0.00 & 100.00 \\
&\textsc{Clarify-and-MA} & -16.18 & 56.66 & 100.00 & 100.00 & -22.25 & 57.77 & 100.00 & 100.00 \\
&\textsc{Answer-FT} & 13.59 & 57.55 & 14.30 & 44.75 & 41.81 & 72.58 & 4.71 & 24.03 \\
\cmidrule(lr){2-10}
&\ourapproach{} & \textbf{54.32} & \textbf{65.73} & 42.86 & 14.60 & \textbf{65.70} & \textbf{79.52} & 20.56 & 9.30 \\
\cmidrule(lr){2-10}
\raisebox{1cm}[0pt]{\begin{sideways}{\textbf{PACIFIC}}\end{sideways}}&\ouroracle{} & 65.24 & 73.89 & 43.46 & 14.05 & 70.95 & 83.60 & 15.60 & 9.39 \\
\bottomrule
\end{tabular}
\medskip
\caption{Performance on \ambigqa{} and \pacific{} development sets using a Gemma 2 9B base model for ambiguous and unambiguous queries, averaged across rollouts. `\% Clar.' and `\% MA' denote the fraction of rollouts with clarification questions and multi-answers. %Fine-tuning on self-play trajectories (\ourapproach) 
\ourapproach{} improves F$_1$ while learning appropriate clarification and multi-answer behavior, leading to overall higher reward.}
\label{tab:results_combined}
\end{table}

\begin{figure}[t] 
    \centering 
   %  \begin{tabular}{@{\hspace{-.2cm}}c@{\hspace{-3.2cm}}c@{}}
   %  \includegraphics[width=.3\textwidth]{figures/alt-coefficient-ambigqa.tex} &
   % \raisebox{-.2cm}[0pt]{\includegraphics[width=.384\textwidth]{figures/alt-coefficient-pacific.tex}}
  %  \includegraphics[width=.85\textwidth]{figures/coefficient_analysis_ambigqa.tex} \\
 %  \raisebox{-3cm}[0pt]{\includegraphics[width=.85\textwidth]{figures/coefficient_analysis_pacific.tex}}
    % \end{tabular}
    % \vspace{-.2cm}
        \vspace{-5pt}
    \includegraphics[width=1.\textwidth]{figures/coefficient_analysis_unified.tex} \\
    \vspace{-5pt}
    \caption{Fraction of rollouts with a clarification question and a multi-answer on the development sets for different cost coefficient values. The $x$-axis for $\alpha$ is in square-root scale and for $\beta$ in log-scale. \ourapproach{} reduces the fraction of clarifications with higher $\alpha$ and reduces the fraction of multi-answers with higher $\beta$, as expected. The baselines do not exhibit the correct sensitivity to coefficient values.\looseness=-1}
    
    % The \label allows you to cross-reference this figure elsewhere in your document using \ref{fig:my_stacked_figure}.
    \label{fig:coeff_analysis}
    \vspace{-10pt}
\end{figure}

\section{Experimental Results}
\label{sec:experimental_results}

\Cref{tab:results_combined} (top) shows results on \ambigqa{}. \ourapproach{} achieves the highest F$_1$ and reward for ambiguous \emph{and} unambiguous queries. \ourapproach{} also dramatically reduces the rate of clarification questions and multi-answers compared to \textsc{Prompted}. Using chain-of-thought improves performance and reduces the rate of clarification questions and multi-answers compared to \textsc{Prompted}, but lags behind \ourapproach{}. The rate of clarification and multi-answer actions in \ourapproach{} is higher for ambiguous queries than for unambiguous queries, as expected.
Training on direct answer observations (\textsc{Answer-FT}) leads to almost always returning a direct answer, but gets higher F$_1$ than \textsc{Answer} since it is finetuned to answer in a better format.
Comparing \ourapproach{} to \ouroracle{}, \ourapproach{} does well in choosing high-reward action sequences, with a similar rate of clarifications and multi-answers. Still, a $\sim$10 point headroom remains.\looseness=-1

\Cref{tab:results_combined} (bottom) shows results on \pacific{}. \ourapproach{} outperforms all baselines in terms of F$_1$ and reward. Unlike \ambigqa{}, \textsc{Prompted} and \textsc{Prompted-COT} use clarification questions and multi-answers more sparingly. Moreover, both methods use more clarifications and multi-answers on ambiguous queries than unambiguous ones, showing they distinguish query types when relevant information is in the prompt. 
\ourapproach{} increases the clarification rate on ambiguous queries to 42.86\%  to improve its performance. \textsc{Prompted-COT} achieves higher reward than \textsc{Prompted}, but trails \ourapproach{} by a large margin. \textsc{Clarify} performs well on ambiguous queries, but worse than \ourapproach{}. As expected, clarification questions and multi-answers occur more frequently in ambiguous queries.
\textsc{Answer-FT} achieves higher F$_1$ compared to \textsc{Answer} as a result of finetuning, but its performance is much lower than \ourapproach{}.
Comparing to \ouroracle{}, we see moderate headroom for \ourapproach{}: 11 points for unambiguous and 5 points for unambiguous queries. Remarkably, the fraction of clarifications and multi-answers for \ourapproach{} closely matches the oracle. 
For ease of reading, we show in~\Cref{tab:results_combined} results without confidence intervals, but we compute and report those in Appendix~\ref{app:confidence_intervals},~\Cref{tab:results_ci}.

\begin{figure}[t] 
    \centering 
    \begin{tabular}{cc}
    \includegraphics[width=0.43\textwidth]{figures/generalization_analysis_ambigqa.tex} &
    \includegraphics[width=0.43\textwidth]{figures/generalization_analysis_pacific.tex}
    \end{tabular}
    \vspace{-.7cm}
    \caption{Generalization experiment: \ourapproach{} on \ambigqa{} and \pacific{} with cost coefficients that did \emph{not} appear during training (purple stars). $x$-axis for~$\alpha$ is in square-root scale and for $\beta$ in log-scale. When varying~$\alpha/\beta$ we fix $\beta=5/\alpha=20$. 
    In both datasets the fraction of clarification questions and multi-answers is monotonic in $\alpha$ and $\beta$, showing good generalization to unobserved cost coefficients.}
    \label{fig:coeff_generalization}
    \vspace{-20pt}
\end{figure}

Since \ourapproach{} is finetuned  on both actions and observations, it learns more than just selecting the right action sequence: for \action{CLARIFY} and \action{MULTI\_ANS}, it learns which clarification questions or multi-answers are likely to yield high reward.
In~\Cref{app:better_clar_example}, we show cases where \ourapproach{} issues better clarification questions than baselines, improving answers.
%\Cref{fig:clarify-sgp-comparison-pacific} shows an example where \ourapproach{} issues a better clarification question compared to \textsc{Prompted}, which eventually leads to a better final answer. A similar example for \ambigqa{} is shown in \Cref{app:better_clar_example}, \Cref{fig:clarify-sgp-comparison}.

We include additional results in the appendix  (\Cref{app:additional_Results}). \Cref{fig:reward_by_coefficient} (\Cref{app:reward_by_coefficient}) shows the average reward for SGP and baselines across all 9 coefficient pairs. The reward improves for \emph{all} coefficient pairs and not just on average. \Cref{fig:act_seq_dist} (\Cref{app:action_seq_distribution}) shows the action distribution across models and datasets, revealing predictable patterns. \Cref{fig:rest_epochs} (\Cref{app:rest_epochs}) shows that reward and F$_1$ improve monotonically across ReST epochs.

% add comment about length

\paragraph{Corpus-level steerability} \Cref{fig:coeff_analysis} shows the fractions of rollouts with clarification questions and with multi-answers for all trained  cost coefficients for \ambigqa{}  and \pacific{} (for brevity, we omit the average length of the final assistant turn, as it behaves similarly to the fraction of multi-answers, but add this metric in App.~\ref{app:length}, \Cref{fig:coeff_analysis_full}). For \ambigqa{}, \textsc{Prompted} behaves counterintuitively, increasing the rate of clarification questions when $\alpha$ increases, and increasing the rate of multi-answers and turn length when $\beta$ increases. \textsc{Prompted-COT} does not change the rate of clarifications or multi-answers as a function of $\alpha$ or $\beta$. \ourapproach{} leads to the desired behavior: increasing $\alpha$ reduces the rate of clarifications, and increasing $\beta$ reduces the rate of multi-answers. In \pacific{}, the behavior of \ourapproach{} is similar to \ambigqa{}, while \textsc{Prompted} and \textsc{Prompted-COT} do not modify their behavior as a function of different coefficients. Unlike \ambigqa{}, \textsc{Prompted-COT} uses clarification questions and multi-answers rarely, while \textsc{Prompted} avoids clarifications, but uses multi-answers. These observations suggest that the base model cannot reason about numerical costs, whereas fine-tuning effectively teaches this capability. 

\begin{wraptable}{t}{0.6\textwidth}
\vspace{-10pt}
\centering
\small
\begin{tabular}{@{}ll@{~} ccc ccc@{}}
\toprule
\multirow{2}{*}{\textbf{Data}} & \multirow{2}{*}{\textbf{Method}} & \multicolumn{3}{c}{$\alpha$: \textbf{0 $\rightarrow$ 20}} & \multicolumn{3}{c}{$\beta$: \textbf{0.1 $\rightarrow$ 5.0}}  \\
\cmidrule(lr){3-5} \cmidrule(lr){6-8} 
 & & R & P & F1 & R & P & F1  \\
\midrule
\multirow{3}{*}{\textsc{Ambig}} & Prom.     & 0.03 & 0.02 & 0.02 & 0.02 & 0.02 & 0.02 \\
 & COT & 0.29 & 0.43 & 0.34 & 0.09 & 0.12 & 0.10 \\
 & \ourapproach{}& 0.45 & 0.49 & 0.47 & 0.09 & 0.42 & 0.12 \\
\midrule
\multirow{3}{*}{\textsc{Pac.}} & Prom. & 0.02 & 0.28 & 0.04 & 0.02 & 0.19 & 0.04 \\
 & COT & 0.03 & 0.21 & 0.05 & 0.01 & 0.09 & 0.02 \\
 & \ourapproach{}& 0.62 & 0.58 & 0.60 & 0.14 & 0.44 & 0.19\\
\bottomrule
\end{tabular}
\caption{\label{tab:prompt-level}  Prompt-level steerability metrics when varying the cost of clarification questions ($\alpha$) and multi-answers ($\beta$). Recall, precision, and F$_1$ are averaged across  three possible values of $\beta$ (clarification questions) and $\alpha$ (multi-answers).}
\vspace{-15pt}
\end{wraptable}
% \end{table}

\paragraph{Prompt-level steerability}
Next, we evaluate steerability at the \emph{prompt level} by asking: for a given prompt, if the cost coefficient changes from one value to another, does the model optimally adjust its action sequence? To answer this, we first identify the \emph{optimal action sequence} for each query and coefficient pair by determining which sequence yields the highest expected reward. We then analyze how the optimal action sequence \emph{changes} as we vary one coefficient while fixing the other. 

For clarification questions, We check if asking (or removing) a clarification question is optimal 
when modifying $\alpha_1 \rightarrow \alpha_2$, while fixing $\beta$. For multi-answers, we check if asking (or removing) a multi-answer is optimal when $\beta_1 \rightarrow \beta_2$, while fixing $\alpha$. 
We then compute recall, precision, and F$_1$ in the following manner. Recall is the fraction of cases where the optimal action sequence changes when changing cost coefficients (e.g., adding a clarification question), and our model predicts the correct change. Precision is the fraction of cases where our model predicts a certain change in the action sequence, and this change is correct according to the optimal action sequence change. F$_1$ is the usual harmonic average.
We describe the details of this procedure in \Cref{app:prompt-level-steerability}.\looseness=-1

Table~\ref{tab:prompt-level} reports results when changing the value of $\alpha$ from 0 $\rightarrow$ 20 and the value of $\beta$ from  0.1 $\rightarrow$ 5.0. \ourapproach{} outperforms baselines across the board. Results are substantially higher for steering $\alpha$ compared to $\beta$, where precision is high but recall is low. We attribute this to the baseline model producing better training rollouts for  asking good clarification questions compared to multi-answers. Lastly, our results show there remains substantial headroom in learning policies that choose the optimal action sequence across prompts.

\textbf{Generalization to new cost coefficients.}
We test if \ourapproach{} can generalize to cost coefficients unobserved at training time. \Cref{fig:coeff_generalization} shows the fraction of clarification questions and multi-answers for observed and unobserved cost coefficients (see purple stars in the figure) for \ambigqa{} (left) and \pacific{} (right). When varying $\alpha$, we fix $\beta=5$, and when varying $\beta$, we fix $\alpha=20$. \ourapproach{} generalizes well to unobserved values of both $\alpha$ and $\beta$ on both \ambigqa{} and \pacific{}, smoothly adjusting the action frequencies between the values corresponding to neighboring coefficients observed at training time. This shows \ourapproach{} yields \emph{generalizable steerability}, giving users control even at cost coefficients that are unobserved during training. Results for the length of the final assistant turn are similar and omitted for brevity.\looseness=-1

\paragraph{Are Gemini models steerable?} 
We saw that Gemma 2 9B can be \emph{trained} to respond to steering signals, but that it is not steerable ``out-of-the-box''. A natural followup is whether stronger models are steerable without fine-tuning. To this end, we compare \ourapproach{} to Gemini-2.5 Flash~\citep{gemini2p5} and Gemini-3 Flash~\citep{gemini3} on \ambigqa{}. 

\Cref{tab:gemini_res} shows the results for this comparison. Gemini models have better parametric knowledge and consequently higher F$_1$. Nevertheless, reward is lower for ambiguous question, and also for unambiguous questions in Gemini-3, due to sub-optimal use of clarification questions and multi-answers. More importantly, Gemini-2.5 is not steerable, with the number of clarification questions increasing (instead of decreasing) when $\alpha$ grows, and no sensitivity to the value of $\beta$. Gemini-3 exhibits moderate steerability, but its overuse of multi-answers reduces its reward. Overall, the performance of even these strong models is sub-optimal when asked to maximize a prompt-specified objective.

\begin{table}[t]
\centering
\scriptsize
\resizebox{\textwidth}{!}{
\begin{tabular}{@{}lrr|rr|rrr|rrr@{}}
\toprule
\multirow{2}{*}{\textbf{Models}} & \multicolumn{2}{c}{\textbf{Ambiguous}} & \multicolumn{2}{c}{\textbf{Unambiguous}} & \multicolumn{3}{c}{\textbf{\% Clarifications ($\alpha$)}} & \multicolumn{3}{c}{\textbf{\% Multi-answers ($\beta$)}} \\
\cmidrule(lr){2-3} \cmidrule(lr){4-5} \cmidrule(lr){6-8} \cmidrule(l){9-11}
& \textbf{Reward} & \textbf{F$_1$} & \textbf{Reward} & \textbf{F$_1$} & \textbf{0.0} & \textbf{2.0} & \textbf{20.0} & \textbf{0.1} & \textbf{0.7} & \textbf{5.0} \\
\midrule
\textsc{Gemini-2.5-Flash} & -15.35 & 23.96 & 51.83 & 60.88 & 23.58 & 34.19 & 49.59 & 24.40 & 21.95 & 25.28 \\
\textsc{Gemini-3-Flash} & -19.93 & 30.55 & 27.95 & 60.40 & 33.86 & 29.75 & 22.42 & 67.43 & 64.91 & 58.10 \\
\midrule
\ourapproach{} & 6.63 & 19.94 & 41.28 & 26.46 & 54.78 & 42.71 & 18.34 & 28.91 & 14.60 & 9.49 \\
\bottomrule
\end{tabular}
}
\medskip
\caption{Performance and steerability of prompted Gemini models compared to \ourapproach{} on \ambigqa{}.}
\label{tab:gemini_res}
\end{table}

\section{Related Work}

\paragraph{Handling ambiguity}
User queries are rife with ambiguities, consistent with the principle of least effort~\citep{zipf1949human}. 
While Grice's \emph{maxim of quantity} suggests that a cooperative speaker provides as much information as necessary~\citep{grice1975logic}, users of AI assistants may assume context that is unavailable to agent, thereby requiring additional grounding. 
Consequently, a growing body of work is focused on handling ambiguity in LLMs~\citep{tanjim2025disambiguation}).
A natural approach is to ask clarification questions~\citep{Dhole2020ResolvingIA,knowno2023,zhang-choi-2025-clarify}: for example, \citet{chen2025learningclarify} and~\citet{zhang2025modelingfutureconversationturns} use self-play with user simulators to learn a policy for asking clarification questions, while  \citet{kim-etal-2024-aligning} let language models assess their own uncertainty and design a corresponding clarification policy.
Other work generates long responses enumerating multiple plausible user interpretations and their answers~\citep{stelmakh-etal-2022-asqa,amplayo-etal-2023-query,kim-etal-2023-tree,in-etal-2025-diversify}.

Recently, \citet{dong-etal-2026-value} proposed an inference-time method for deciding when to ask clarification questions that is based on explicit estimation of the expected value of each possible agent action. 
Our \textsc{Prompted-COT} baseline takes a similar approach, but makes less assumptions on the space of possible clarification questions and answers.

\paragraph{Steerable language policies}
Teaching contextual steerability has been a north star of language generation~\citep{hovy1987some,hovy1988two,chen2022seamlessly,keskarCTRL2019,sorensen2025spectrumtuningposttrainingdistributional}. 
Recent work on adapting to uncertainty includes \citet{cole-etal-2023-selectively}, who teach models to abstain from answering questions when uncertain about the asker's intent, and \citet{kuhn2023clamselectiveclarificationambiguous}, who prompt models to quantify their uncertainty to determine when to ask clarification questions.
However, these approaches are ``steerable'' only by the level of uncertainty, not by other contextual factors.
\citet{wang-etal-2024-conditional} train \emph{conditional language policies} (CLPs) for steerable alignment: given multiple reward models, they train to maximize a linear combination of rewards, where the combination coefficients are sampled during training and provided as input at test time. We use a similar approach for training steerable grounding policies under different cost considerations. However, our approach uses coefficients specified in the prompts, allowing for training a \emph{single} model, whereas \citet{wang-etal-2024-conditional} train \emph{multiple} models (with some parameter sharing) to achieve steerability.\looseness=-1
%However, \citet{wang-etal-2024-conditional} found that conditioning on prompt-inserted coefficients is ineffective, and were therefore forced to train \emph{multiple policies} (with some parameter sharing) to obtain steerability -- their experiments with prompting showed limited steerability. We show that collaborative self-play on a Gemma backbone~\cite{gemmateam2024gemma2improvingopen} makes it possible to learn steerability from prompted reward coefficients. Another distinction is that we focus on steering with interpretable parameters of the interaction style (brevity and interactivity), whereas \citet{wang-etal-2024-conditional} steer between more opaque evaluation metrics.

Finally, \citet{sun2025trainingproactivepersonalizedllm} use RL to train models to ask clarification questions and respond to prompt-specified user prompts for agentic tasks. In contrast, we focus on training assistants to maximize a reward function specified by \emph{numerical cost coefficients}. In addition, we closely link steerability to the use of clarification, viewing steerability as a knob for controlling the spectrum of possible grounding strategies. In contrast, \citet{sun2025trainingproactivepersonalizedllm} focus on steerability for stylistic personalization, which they treat as orthogonal to posing clarification questions.

\paragraph{Multi-agent self-play}
A large body of work explores the use of multi-agent interaction as part of post-training (in contrast to inference-time multi-agent systems) for improving reasoning~\citep{chen2024magdi,subramaniam2025multiagent} and factuality~\citep{khan2024debate}. Other lines of research (including this work) focus on self-play for improving the collaborative and pragmatic skills of assistants~\citep{lin2024decision}. Beyond ambiguity, past work has also explored self-play for improving uncertainty expression~\citep{stengel-eskin2024lacie,eisenstein2025dontliefriendslearning}, tool use~\citep{eisenstein2025dontliefriendslearning}, handling of persuasion~\citep{stengel-eskin-etal-2025-teaching}, personalization~\citep{wagner2024personalized}, and general collaborative skills~\citep{wu2025collabllm}. Our work uses multi-agent self play as a training framework, applying it to the objective of learning a 
cost-sensitive, dynamically steerable policy for managing uncertainty. 

\vspace{-5pt}
\section{Conclusions}
\label{sec:conlusions}

\vspace{-5pt}

\begin{comment} % a bit repetitive, see my alternative below -J
This work focuses on managing the uncertainty of AI assistants when interacting with users. 
We argue that disambiguation strategies are highly contextual, and therefore propose to train models that choose the right strategy at the right time.
To achieve this, we provide the AI assistants with a parameterized objective in its prompt, which specifies the cost of each strategy. We then use reinforced self-training to train the base model to change its behavior given different cost parameters in order to maximize this objective. 
We evaluate our approach on two question answering benchmarks where some of the input queries are ambiguous (either with respect to a particular context or not), and show that our approach is effective -- it leads to models that properly condition on provided cost coefficients to maximize the reward provided in their prompt, and even generalize to coefficients that did not appear at training time.
\end{comment}

Prior work on uncertainty has focused on hedging and clarification in isolation, addressing \emph{single} disambiguation strategies. We argue that choosing the \emph{right} disambiguation strategy is  contextual, and propose training models that select a cost-optimal strategy, based on numerical steering coefficients. We achieve this through \ourapproach, yielding cost-accuracy tradeoffs that are responsive to steering signals while also being pareto-optimal in aggregate. %\ourapproach-trained models even generalize to steering coefficients unseen during training.
Looking forward, a natural next step is to close the loop and infer user preferences over time, leading to proactive and flexible AI assistants that infer user preferences and adapt to them on-the-fly.
We focused on three disambiguation strategies --- guessing, clarifying, and hedging --- but the range of subtle grounding actions, that human speakers use to signal interpretation difficulties is much wider, often including nonverbal cues.
Extending \ourapproach{} to such grounding actions is a challenge for future work.
Finally, steering models with numerical values has applications beyond uncertainty management, in areas such as controlling tool use, test-time compute, and more.

\section*{Acknowledgments}
We thank Alekh Agarwal, William Cohen, Chris Dyer, Tal Linzen, and Eric Malmi for their thoughtful feedback. We thank  Liu Leqi for discussions on training data generation. We thank Dheeru Dua and Vicky Zayats for developing supporting infrastructure.

\bibliography{colm2026_conference}
\bibliographystyle{colm2026_conference}
\clearpage
\appendix

\section{LLM Disclosure}
We used language and image models in the following ways: (1) creating figures and formatting tables by generating LaTeX code; (2) generating Figures~\ref{fig:intro}, \ref{fig:rollout}, and \ref{fig:nonparametric_knowledge}; (3) writing data analysis code; (4) reviewing the paper and suggesting areas in which the writing could be improved.

\section{Additional Details on the \pacific{} Dataset}
\label{app:pacific_examples}

\Cref{fig:nonparametric_knowledge} shows an illustrative example from \pacific{}, where the query  is ambiguous given the provided table context, which is shown only to the assistant.

\begin{figure}[h!]
    \centering
    \includegraphics[width=0.6\linewidth]{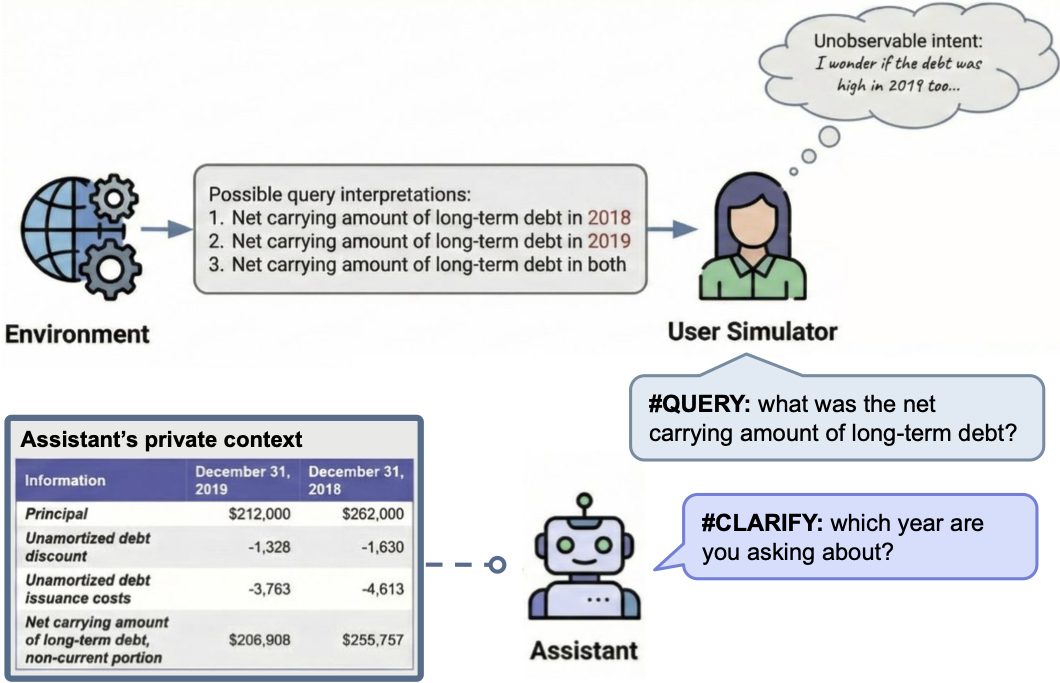}
    \caption{The assistant has access to a private context (table in this case), which 
    is necessary for determining ambiguity of the query. The user does not have access to this context.}
    \label{fig:nonparametric_knowledge}
\end{figure}

\Cref{fig:multiturn-ambiguity} depicts an example of the different types of conversation phenomena that occur in the \pacific ~corpus. In \ambigqa, all questions are inherently single-turn, and the source of whether it is ambiguous or not has to come from parametric knowledge. In contrast, as highlighted in the \textcolor{blue}{blue} text of \autoref{fig:multiturn-ambiguity}, we can see that the source of ambiguity in the question ``What was the Net carrying amount of long-term debt?'' comes from the contextual knowledge --- the Assistant is provided a table which contains two different possible years with ``Net carrying amount of long-term debt.'' We can also see by the text highlighted in \textcolor{red}{red} that although a single-turn question such as ``How about 2018?'' may be underspecified, the question is not ambiguous due to the multi-turn context which indicates that the user is still asking about ``Net carrying amount of long-term debt.''

\begin{figure}[H]
    \centering
    \footnotesize 
    \begin{minipage}[t]{0.65\textwidth}
        \renewcommand{\arraystretch}{1.3} 
        \begin{tabular}{@{}lrr@{}} 
            \toprule
            & \multicolumn{2}{c}{\textbf{Data}} \\
            \cmidrule(l){2-3}
            & \textbf{Dec 31, 2019} & \textbf{Dec 31, 2018} \\
            \midrule
            Principal & \$212,000 & \$262,000 \\
            \textit{Less} & & \\
            Unamortized debt discount & -1,328 & -1,630 \\
            Unamortized debt issuance costs & -3,763 & -4,613 \\
            \cmidrule(l){2-3} 
            Net carrying amount & 206,909 & 255,757 \\
            \multicolumn{3}{l}{Less: current portion of long-term debt} \\
            Long-term debt, non-current portion & \textbf{\$206,909} & \textbf{\$255,757} \\
            \bottomrule
        \end{tabular}
    \end{minipage}
    \hfill 
    \begin{minipage}[t]{0.30\textwidth}
        \setlength{\parindent}{0pt}
        \setlength{\parskip}{0.5em}
        \raggedright
        \vspace{-2.9cm} 
        
        \textbf{User:} What was the \textcolor{chatblue}{Net carrying amount of long-term debt?}
        
        \textbf{Assistant:} \textcolor{chatblue}{What year are you asking about?}
        
        \textbf{User:} 2019
        
        \textbf{Assistant:} 206,909 thousand
        
        \textbf{User:} \textcolor{chatred}{How about 2018?}
        
        \textbf{Assistant:} 255,757 thousand
        
        \textbf{User:} In which year was \textcolor{chatred}{that amount} less than 210,000 thousands?
        
        \textbf{Assistant:} 2019
    \end{minipage}
    \caption{
    \textbf{Conversational phenomena in \pacific.} 
        \textcolor{chatblue}{Blue text}: ambiguity occurs due to contextual grounding on the input table. 
        \textcolor{chatred}{Red text}: single-turn underspecification does not imply ambiguity given multi-turn context.
    }
    \label{fig:multiturn-ambiguity}
\end{figure}

\section{Examples for Improved Clarification Questions}
\label{app:better_clar_example}

\Cref{fig:clarify-sgp-comparison-pacific} and~\Cref{fig:clarify-sgp-comparison} show examples from \pacific{} and \ambigqa{} where \ourapproach{} learns to issue a good clarification question, which leads to a more accurate answer.

\definecolor{bg_gray}{RGB}{245,245,245}
\begin{figure}[h]
    \centering
    \begin{tcolorbox}[colback=bg_gray, colframe=black!70]
        \small
        \textbf{Context:} \emph{...[Table and initial conversation omitted for brevity]...} \\
        \textbf{Query:} In which year was \textbf{Transition costs and project assets} less than 100 million? \\
        \textbf{Intent:} In which year was \textbf{Transition costs and project assets} under \textbf{other current assets} less than 100 million? \\
        \textbf{Gold:} 2020
        \tcbline
        \begin{minipage}[t]{0.48\linewidth}
            \textbf{\underline{\ourapproach{}}} \par \vspace{3pt}
            \textbf{Assistant:} \action{CLARIFY}: Which category of Transition costs and project assets ? \\
            \textbf{User:} \action{RESPOND}: under other current assets \\
            \textbf{Assistant:} \action{ANSWER}: 2020
        \end{minipage}
        \hfill
        \rule[-2cm]{0.5pt}{2.2cm}
        % \rule[-3cm]{0.5pt}{3cm} % Vertical divider line
        \hfill
        \begin{minipage}[t]{0.48\linewidth}
            \textbf{\underline{\textsc{Prompted}}} \par \vspace{3pt}
            \textbf{Assistant:} \action{CLARIFY}: What do you mean by less than 100 million? \\
            \textbf{User:} \action{RESPOND}: Under 100 million \\
            \textbf{Assistant:} \action{ANSWER}: 2018
        \end{minipage}
    \end{tcolorbox}
    \caption{A \pacific{} example where \ourapproach{} asks a clarification question that leads to higher reward compared to the \textsc{Prompted} baseline. Context is omitted for brevity; column names from the table are in bold. The table contains the column \textbf{Transition costs and project assets} twice: once under the category 
    \textbf{Other Current Assets} and once under the category \textbf{Other Assets}. The answer is ``2020'' for the correct category, and ``2018'' for the wrong one.}
    \label{fig:clarify-sgp-comparison-pacific}
\end{figure}

\definecolor{bg_gray}{RGB}{245,245,245}
\begin{figure}[h]
    \centering
    \begin{tcolorbox}[colback=bg_gray, colframe=black!70]
        \small
        \textbf{Query:} When did frontier take over verizon in california? \\
        \textbf{Intent:} When did frontier officially take over of all of verizon's wired telecom operations in california? \\
        \textbf{Gold:} April 1, 2016
        \tcbline
        \begin{minipage}[t]{0.48\linewidth}
            \textbf{\underline{\ourapproach{}}} \par \vspace{3pt}
            \textbf{Assistant:} \action{CLARIFY}: Do you mean when Frontier Communications acquired Verizon's landline business in California, or are you asking about a different type of takeover? \\
            \textbf{User:} \action{RESPOND}: Frontier Communications acquired Verizon's landline business in California. \\
            \textbf{Assistant:} \action{ANSWER}: April 1, 2017
        \end{minipage}
        \hfill
        \rule[-3cm]{0.5pt}{3cm} % Vertical divider line
        \hfill
        \begin{minipage}[t]{0.48\linewidth}
            \textbf{\underline{\textsc{Prompted}}} \par \vspace{3pt}
            \textbf{Assistant:} \action{CLARIFY}: Do you mean a specific year or just a general timeframe (e.g., 2000s)? \\
            \textbf{User:} \action{RESPOND}: a specific year \\
            \textbf{Assistant:} \action{ANSWER}: 2009
        \end{minipage}
    \end{tcolorbox}
    \caption{An \ambigqa{} example where \ourapproach{} asks a clarification question that leads to higher reward compared to \textsc{Prompted}.}
    \label{fig:clarify-sgp-comparison}
\end{figure}

\section{Ablations on the Training Procedure}
\label{app:ablations}

Our training algorithm is a slight variant of ReST~\citep{gulcehre2023reinforced}, as we describe in~\Cref{sec:method}. We conduct two ablations on this training procedure. First, we check whether \emph{grouping} of rollouts based on action sequences is necessary by ablating this part and simply choosing the highest-reward rollout per query without any grouping. Second, we check if training on observations is beneficial by training on actions only. For both ablations we conduct a single round of ReST.

\Cref{tab:pacific_ablation} and~\Cref{tab:ambigqa_ablation} show the results of these ablations. We observe that in almost all cases our approach outperforms the ablations, typically by a large margin.

\begin{table}[hbt!]
\centering
\begin{tabular}{lcccc}
\toprule
\textbf{Models} & \textbf{Ambig. Reward} & \textbf{Ambig. F$_1$} & \textbf{Unambig. Reward} & \textbf{Unambig. F$_1$} \\
\midrule
Prompted & 11.46 & 55.28 & 42.44 & 69.51 \\
ReST-1 & 45.75 & 62.39 & 58.57 & 76.49 \\
\midrule
No Grouping & 44.80 & 61.24 & 56.12 & 76.04 \\
Action Only & 33.10 & 50.79 & 42.81 & 62.96 \\
\bottomrule
\end{tabular}
\caption{Ablation results on \pacific{}, where we train for one round of ReST and compare our approach (ReST-1) with variants where (a) no rollout grouping is performend, and (b) we train on actions only, but not on observations.}
\label{tab:pacific_ablation}
\end{table}

\begin{table}[hbt!]
\centering
\begin{tabular}{lcccc}
\toprule
\textbf{Models} & \textbf{Ambig. Reward} & \textbf{Ambig. F$_1$} & \textbf{Unambig. Reward} & \textbf{Unambig. F$_1$} \\
\midrule
Prompted & -34.49 & 16.16 & -15.01 & 33.07 \\
Rest-1 & 2.03 & 17.26 & 26.30 & 38.38 \\
\midrule
No Grouping & -11.54 & 18.35 & 9.31 & 37.75 \\
Action Only & 4.46 & 6.82 & 7.82 & 9.60 \\
\bottomrule
\end{tabular}
\caption{Ablation results on \ambigqa{}, where we train for one round of ReST and compare our approach (ReST-1) with variants where (a) no rollout grouping is performend, and (b) we train on actions only, but not on observations.}
\label{tab:ambigqa_ablation}
\end{table}

\section{Statistics on Rollout Filtering}
\label{app:filter_stats}

As explained in~\Cref{sec:experimental_results},  when generating training data we filter rollouts with low F$_1$ or when the user simulator ignores the assistant. \Cref{tab:ambigqa_filter_stats} and~\Cref{tab:pacific_filter_stats} provide statistics on how often this filtering happen across ReST epochs.
As evident, the fraction of filtered rollouts goes down during training.

\begin{table}[hbt!]
\centering
\begin{tabular}{lcc}
\toprule
\textbf{Training stage} & \textbf{$F_1 <$ threshold} & \textbf{Simulator violation} \\
\midrule
ReST-1 & 66\% & 5\% \\
ReST-2 & 57\% & 3\% \\
ReST-3 & 52\% & 3\% \\
\bottomrule
\end{tabular}
\caption{Fraction of rollouts that is filtered across ReST epochs in \ambigqa.}
\label{tab:ambigqa_filter_stats}
\end{table}

\begin{table}[hbt!]
\centering
\begin{tabular}{lcc}
\toprule
\textbf{Training stage} & \textbf{$F_1 <$ threshold} & \textbf{Simulator violation} \\
\midrule
ReST-1 & 25\% & 4\% \\
ReST-2 & 14\% & 3\% \\
ReST-3 & 11\% & 3\% \\
\bottomrule
\end{tabular}
\caption{Fraction of rollouts that is filtered across ReST epochs in \pacific.}
\label{tab:pacific_filter_stats}
\end{table}

\section{Confidence Intervals for Main Results}
\label{app:confidence_intervals}

\Cref{tab:results_ci} shows 95\% confidence intervals computed with bootstrap sampling for the main results presented in~\Cref{tab:results_combined}. As can be seen, confidence intervals are narrow, illustrating the significance of our results.

\begin{table}[hbt!]
\centering
\scriptsize
\resizebox{\textwidth}{!}{
\begin{tabular}{@{}ll|l>{\columncolor{gray!20}}l>{\columncolor{gray!20}}l>{\columncolor{gray!20}}l|l>{\columncolor{gray!20}}l>{\columncolor{gray!20}}l>{\columncolor{gray!20}}l@{}}
\toprule
& \multirow{2}{*}{\textbf{Models}} & \multicolumn{4}{c}{\textbf{Ambiguous}} & \multicolumn{4}{c}{\textbf{Unambiguous}} \\
\cmidrule(lr){3-6} \cmidrule(lr){7-10}
& & \textbf{Reward} & \textbf{F1} & \textbf{\% Clar.} & \textbf{ \% MA} & \textbf{Reward} & \textbf{F1} & \textbf{\% Clar.} & \textbf{\% MA} \\
\midrule
%\multicolumn{9}{@{}l}{\textbf{AmbigQA}} \\
%\midrule
&\textsc{Prompted} & -34.5 [-35.1, -33.9] & 16.2 [15.9, 16.4] & 66.8 [66.5, 67.2] & 63.9 [63.5, 64.3] & -15.0 [-16.7, -13.2] & 33.1 [32.1, 34.0] & 64.8 [63.7, 65.8] & 58.1 [57.0, 59.2] \\
&\textsc{Prompted-COT} & -20.1 [-20.7, -19.5] & 16.4 [16.2, 16.7] & 53.9 [53.5, 54.3] & 50.5 [50.1, 50.9] & -5.0 [-6.7, -3.3] & 29.1 [28.3, 30.0] & 48.3 [47.2, 49.4] & 49.6 [48.4, 50.7] \\
\cmidrule(lr){2-10}
&\textsc{Answer} & 1.3 [1.0, 1.7] & 14.3 [14.1, 14.6] & 0.0 [0.0, 0.0] & 0.0 [0.0, 0.0] & 26.7 [25.7, 27.8] & 35.6 [34.6, 36.5] & 0.0 [0.0, 0.0] & 0.0 [0.0, 0.0] \\
&\textsc{Clarify} & -9.6 [-10.0, -9.2] & 16.3 [16.0, 16.6] & 100.0 [100.0, 100.0] & 0.0 [0.0, 0.0] & 8.8 [7.7, 10.0] & 31.9 [31.0, 32.7] & 100.0 [100.0, 100.0] & 0.0 [0.0, 0.0] \\
&\textsc{Multi-answer (MA)} & -46.5 [-47.2, -45.8] & 15.7 [15.4, 15.9] & 0.0 [0.0, 0.0] & 100.0 [100.0, 100.0] & -25.4 [-27.2, -23.6] & 33.8 [32.8, 34.7] & 0.0 [0.0, 0.0] & 100.0 [100.0, 100.0] \\
&\textsc{Clarify-and-MA} & -50.6 [-51.2, -50.0] & 16.9 [16.7, 17.2] & 100.0 [100.0, 100.0] & 100.0 [100.0, 100.0] & -35.3 [-37.1, -33.5] & 32.3 [31.3, 33.2] & 100.0 [100.0, 100.0] & 100.0 [100.0, 100.0] \\
&\textsc{Answer-FT} & 3.6 [3.3, 4.0] & 16.4 [16.1, 16.7] & 8.7 [8.5, 9.0] & 10.0 [9.8, 10.2] & 31.0 [29.8, 32.1] & 39.6 [38.7, 40.6] & 2.3 [2.0, 2.6] & 4.6 [4.1, 5.1] \\
\cmidrule(lr){2-10}
\raisebox{1cm}[0pt]{\begin{sideways}{\textbf{AmbigQA}}\end{sideways}}&\ourapproach{} & \textbf{6.6 [6.3, 7.0]} & \textbf{19.9 [19.7, 20.2]} & 40.2 [39.8, 40.7] & 18.2 [17.9, 18.5] & \textbf{31.3 [30.2, 32.5]} & \textbf{41.3 [40.3, 42.2]} & 26.5 [25.5, 27.5] & 14.0 [13.2, 14.8] \\ 
\midrule
\midrule
% \multicolumn{10}{@{}l}{} \\ \midrule
&\textsc{Prompted} & 11.5 [11.0, 12.0] & 55.3 [55.0, 55.5] & 18.0 [17.7, 18.2] & 47.2 [46.9, 47.5] & 42.4 [41.7, 43.2] & 69.5 [69.2, 69.9] & 4.8 [4.6, 5.0] & 19.1 [18.7, 19.6] \\
&\textsc{Prompted-COT} & 33.0 [32.4, 33.6] & 50.0 [49.7, 50.4] & 16.5 [16.0, 16.9] & 16.2 [15.8, 16.6] & 52.1 [51.1, 53.0] & 68.4 [67.8, 69.1] & 4.9 [4.5, 5.3] & 7.2 [6.7, 7.6] \\
\cmidrule(lr){2-10}
&\textsc{Answer} & 42.4 [42.2, 42.7] & 52.3 [52.0, 52.5] & 0.0 [0.0, 0.0] & 0.0 [0.0, 0.0] & 56.9 [56.5, 57.4] & 71.5 [71.2, 71.9] & 0.0 [0.0, 0.0] & 0.0 [0.0, 0.0] \\
&\textsc{Clarify} & 45.5 [45.0, 46.0] & 62.7 [62.3, 63.1] & 100.0 [100.0, 100.0] & 0.0 [0.0, 0.0] & 46.9 [46.0, 47.8] & 68.9 [68.2, 69.5] & 100.0 [100.0, 100.0] & 0.0 [0.0, 0.0] \\
&\textsc{Multi-answer (MA)} & -21.4 [-22.5, -20.3] & 54.6 [54.2, 55.0] & 0.0 [0.0, 0.0] & 100.0 [100.0, 100.0] & -4.1 [-5.8, -2.3] & 63.3 [62.7, 64.0] & 0.0 [0.0, 0.0] & 100.0 [100.0, 100.0] \\
&\textsc{Clarify-and-MA} & -16.2 [-17.2, -15.1] & 56.7 [56.3, 57.0] & 100.0 [100.0, 100.0] & 100.0 [100.0, 100.0] & -22.2 [-24.1, -20.4] & 57.8 [57.2, 58.5] & 100.0 [100.0, 100.0] & 100.0 [100.0, 100.0] \\
&\textsc{Answer-FT} & 13.6 [12.6, 14.5] & 57.6 [57.2, 57.9] & 14.3 [13.9, 14.7] & 44.7 [44.2, 45.3] & 41.8 [40.4, 43.2] & 72.6 [72.0, 73.1] & 4.7 [4.3, 5.1] & 24.0 [23.2, 24.9] \\
\cmidrule(lr){2-10}
\raisebox{1cm}[0pt]{\begin{sideways}{\textbf{PACIFIC}}\end{sideways}}&\ourapproach{} & \textbf{54.3 [53.9, 54.8]} & \textbf{65.7 [65.3, 66.1]} & 42.9 [42.3, 43.4] & 14.6 [14.2, 15.0] & \textbf{65.7 [65.0, 66.4]} & \textbf{79.5 [79.0, 80.0]} & 20.6 [19.9, 21.3] & 9.3 [8.7, 9.8] \\
\bottomrule
\end{tabular}
}
\medskip
\caption{Performance on \ambigqa{} and \pacific{} development sets with $95\%$ confidence intervals. `\% Clar.' and `\% MA' denote the fraction of rollouts with clarification questions and multi-answers.}
\label{tab:results_ci}
\end{table}

\section{Additional Empirical Results and Analysis}
\label{app:additional_Results}

\subsection{Results with Gemma 2 2B}
\label{app:gemma_2_2b}

Our main results in~\Cref{sec:experimental_results} use Gemma 2 9B. Here, we present additional experiments using Gemma 2 2B. Since the parameteric knowledge of Gemma 2 2B is weak compared to Gemma 2 9B, we restrict our experiments to \pacific{}, which does not rely on parametric knowledge. In addition, we perfom only a single round of ReST. Clearly, SGP improves over the baselines.

\begin{table}[hbt!]
\centering
\scriptsize
\begin{tabular}{@{}llrr|rr@{}}
\toprule
& \multirow{2}{*}{\textbf{Models}} & \multicolumn{2}{c}{\textbf{Ambiguous}} & \multicolumn{2}{c}{\textbf{Unambiguous}} \\
\cmidrule(lr){3-4} \cmidrule(lr){5-6}
& & \textbf{Reward} & \textbf{F1} & \textbf{Reward} & \textbf{F1} \\
\midrule
&\textsc{Prompted} & 26.27 & 39.13 & 37.57 & 55.76 \\
&\textsc{Prompted-COT} & 21.12 & 34.39 & 29.68 & 48.26 \\
&\textsc{SGP-ReST-1} & 43.34 & 53.67 & 51.89 & 68.19 \\
\bottomrule
\end{tabular}
\medskip
\caption{Performance on the \pacific{} development set for selected models.}
\label{tab:pacific_gemma_2_2b}
\end{table}

In addition, we can evaluate corpus-level steerability of Gemma 2 2B by computing the fraction of rollouts with clarification questions as we vary $\alpha$ and the fraction of rollouts with multi-answers as we vary $\beta$. \Cref{tab:coefficient_sensitivity_gemma2_2b} shows that indeed \ourapproach{} behaves as expected, while the baselines are not steered by the cost coefficients.

\begin{table}[hbt!]
\centering
\scriptsize
\begin{tabular}{@{}lrrr|rrr@{}}
\toprule
\multirow{2}{*}{\textbf{Models}} & \multicolumn{3}{c}{\textbf{\% Clarifications ($\alpha$)}} & \multicolumn{3}{c}{\textbf{\% Multi-answers ($\beta$)}} \\
\cmidrule(lr){2-4} \cmidrule(l){5-7}
& \textbf{0.0} & \textbf{2.0} & \textbf{20.0} & \textbf{0.1} & \textbf{0.7} & \textbf{5.0} \\
\midrule
\textsc{Prompted} & 19.77 & 17.85 & 17.43 & 19.39 & 18.70 & 18.78 \\
\textsc{Prompted-COT} & 36.78 & 36.07 & 35.50 & 18.89 & 19.06 & 18.91 \\
\ourapproach{} & 35.37 & 29.41 & 12.23 & 30.34 & 24.77 & 12.56 \\
\bottomrule
\end{tabular}
\medskip
\caption{Fraction of rollouts with a clarification question and a multi-answer across different cost coefficient values with Gemma 2 2B. As expected, \ourapproach{} reduces the fraction of clarifications as $\alpha$ increases, and reduces multi-answers as $\beta$ increases, whereas baselines remain mostly static.}
\label{tab:coefficient_sensitivity_gemma2_2b}
\end{table}

\subsection{Results with Qwen 3 4B}
\label{app:qwen_3_4b}

We additionally conducted experiments on a separate model family altogether --- Qwen 3 4B. A noticeable difference between Qwen 3 4B and the Gemma 2 family is that Qwen 3 is a \textit{thinking-enabled model}. To adjust for this difference, during rollouts, we still sample from the policy model and evaluate the observed responses, but we include the elicited thinking traces in the data accumulated throughout ReST. We still observe that SGP improves over the baseline, with particularly large reward improvements.

\begin{table}[hbt!]
\centering
\scriptsize
\begin{tabular}{@{}llrr|rr@{}}
\toprule
& \multirow{2}{*}{\textbf{Models}} & \multicolumn{2}{c}{\textbf{Ambiguous}} & \multicolumn{2}{c}{\textbf{Unambiguous}} \\
\cmidrule(lr){3-4} \cmidrule(lr){5-6}
& & \textbf{Reward} & \textbf{F1} & \textbf{Reward} & \textbf{F1} \\
\midrule
&\textsc{Prompted} & -265.92 & 27.61 & -82.63 & 42.01 \\
&\textsc{SGP} & 3.83 & 38.10 & 20.01 & 48.44 \\
\bottomrule
\end{tabular}
\medskip
\caption{Performance on the \pacific{} development set for 
\ourapproach{} compare to 
textsc{Prompted}.}
\label{tab:pacific_qwen_3_4b}
\end{table}

We also evaluated corpus-level steerability of Qwen 3 4B by computing the fraction of rollouts with clarification questions as we vary $\alpha$ and the fraction of rollouts with multi-answers as we vary $\beta$. \Cref{tab:coefficient_sensitivity_qwen3_4b} shows that whereas the baseline prompted model shows no clear trend, \ourapproach{} behaves as expected with consistent decreases in the percentage of clarifying questions as $\alpha$ increases and consistent decreases in the percentage of multi-answers as $\beta$ decreases.

\begin{table}[hbt!]
\centering
\scriptsize
\begin{tabular}{@{}lrrr|rrr@{}}
\toprule
\multirow{2}{*}{\textbf{Models}} & \multicolumn{3}{c}{\textbf{\% Clarifications ($\alpha$)}} & \multicolumn{3}{c}{\textbf{\% Multi-answers ($\beta$)}} \\
\cmidrule(lr){2-4} \cmidrule(l){5-7}
& \textbf{0.0} & \textbf{2.0} & \textbf{20.0} & \textbf{0.1} & \textbf{0.7} & \textbf{5.0} \\
\midrule
\textsc{Prompted} & 3.42 & 3.29 & 3.56 & 15.85 & 15.89 & 16.14 \\
\ourapproach{} & 6.34 & 4.85 & 2.53 & 16.26 & 14.89 & 10.44 \\
\bottomrule
\end{tabular}
\medskip
\caption{Fraction of rollouts with a clarification question and a multi-answer across different cost coefficient values with Qwen 3 4B. As expected, \ourapproach{} reduces the fraction of clarifications as $\alpha$ increases, and reduces multi-answers as $\beta$ increases, whereas baselines remain mostly static.}
\label{tab:coefficient_sensitivity_qwen3_4b}
\end{table}

\subsection{Analysis of average reward per coefficient-pair value}
\label{app:reward_by_coefficient}

Figure~\ref{fig:reward_by_coefficient} shows the average reward for Prompted, Prompted-COT and \ourapproach{} for all values of cost coefficient pairs. The average reward improves across all values and not only on average.

\begin{figure}[h]
    \centering
    \includegraphics[width=0.49\textwidth]{figures/reward_by_coefficient_ambigqa.tex}    
    \includegraphics[width=0.49\textwidth]{figures/reward_by_coefficient_pacific.tex}    
    \caption{Average reward per coefficient for \ambigqa{} (left) and \pacific{} (right). \ourapproach{} improves reward uniformly across all cost coefficient values. In \ambigqa{} when $\beta$ is high, rewards can be negative due to the high cost of each generated word.}
    \label{fig:reward_by_coefficient}
\end{figure}

\subsection{Action sequence distribution}
\label{app:action_seq_distribution}

\begin{figure*}[t] 
    \centering 
    \includegraphics[width=0.49\textwidth]{figures/act_seq_distribution_ambigqa_tikz.tex}    
    \includegraphics[width=0.49\textwidth]{figures/act_seq_distribution_pacific_tikz.tex}
    \caption{Action sequence distributions for \textsc{Prompted}, \textsc{Prompted-COT}, and \ourapproach{} for ambiguous and unambiguous queries in \ambigqa{} (left) and \pacific{} (right) development sets. Action sequences are shown using their first letter: Q: \action{QUERY}, A: \action{ANSWER}, C: \action{CLARIFY}, R: \action{RESPOND}, M: \action{MULTI\_ANS}, F: \action{FINALIZE} (see \Cref{tab:agent_actions}) In both \ambigqa{} and \pacific{} we see that the fraction of direct answers (without clarification or multi-answer) is higher in unambiguous queries compared to ambiguous ones.
    }
    \label{fig:act_seq_dist}
\end{figure*}

\Cref{fig:act_seq_dist} shows the action distribution for \textsc{Prompted}, \textsc{Prompted-COT}, and \ourapproach{} for ambiguous and unambiguous queries in \ambigqa{} (left) and \pacific{} (right). In \ambigqa{}, the prompted model tends to overuse clarification questions and multi-answers. \textsc{Prompted-COT} has a more uniform distribution across action sequences. Finally, \ourapproach{}  answers directly more frequently, and as expected uses more clarification questions for ambiguous queries.

In \pacific{}, \textsc{Prompted} and \textsc{Prompted-COT} often answer directly. This may be because the model infers that it can already answer successfully given the conversation history and table. However, these prompted approaches do use more direct answers for unambiguous queries compared to ambiguous ones, showing some understanding of ambiguity.
Fine-tuning the model with \ourapproach{} changes the distribution, which becomes more similar to \ambigqa{}, where answering directly is the most common action sequence, but clarification questions are also quite common, especially when the query is ambiguous.

\subsection{Improvements across ReST iterations}
\label{app:rest_epochs}

\begin{figure*}[t] 
    \centering 
    \includegraphics[width=1.0\textwidth]{figures/rest_epochs.tex}    
    \caption{Results on the development set across ReST epochs for both \ambigqa{} and \pacific{}.}
    
    % The \label allows you to cross-reference this figure elsewhere in your document using \ref{fig:my_stacked_figure}.
    \label{fig:rest_epochs}
\end{figure*}

\Cref{fig:rest_epochs} shows the F$_1$ and reward for ambiguous and unambiguous queries on the development sets of \ambigqa{} and \pacific{}. The table shows that metrics monotonically improve throughout iterations. An exception is the third ReST epoch of \pacific{}, where results are slightly lower than the second epoch, which we attribute to some training instability.
For \ambigqa{} additional iterations might potentially further improve performance.

\subsection{Average length of final assistant turn across baselines}
\label{app:length}

\Cref{fig:coeff_analysis_full} shows the same results as \Cref{fig:coeff_analysis}, but with an additional metric, the average length of the final answer provided by the assistant, a quantity that goes into reward computation. Since the trends are very similar to the fraction of multi-answers, this was omitted from the main text.

\begin{figure}[h] 
    \centering 
    \includegraphics[width=0.9\textwidth]{figures/coefficient_analysis_ambigqa.tex}
    \includegraphics[width=0.9\textwidth]{figures/coefficient_analysis_pacific.tex}
    \caption{Fraction of rollouts with a clarification question (left), fraction of rollouts with a multi-answer (middle), and average length of the final assistant answer (right) for Prompted, Prompted-COT, and \ourapproach{} on the development sets of \ambigqa{} (top) and \pacific{} (bottom) for different values of cost coefficients. The $x$-axis for $\alpha$ is in square-root scale and for $\beta$ in logarithmic scale. In both datasets, \ourapproach{} reduces the fraction of clarifications with higher $\alpha$ and reduces the fraction of multi-answers with higher $\beta$, as expected. \textsc{Prompted} and \textsc{Prompted-COT} do not exhibit this correct sensitivity to values of cost coefficients.}
    \label{fig:coeff_analysis_full}
\end{figure}

\section{Prompt-level Steerability Metric}
\label{app:prompt-level-steerability}

Here, we describe the steps for computing the prompt-level steerability metric shown in~\Cref{sec:experimental_results} and~\Cref{tab:prompt-level}.
Let's assume we are computing prompt-level steerability when modifying $\alpha_1 \rightarrow \alpha_2$ ($\beta_1 \rightarrow \beta_2$ is similar).

\begin{enumerate}
    \item For each query $q$, cost-coefficients $\alpha, \beta$, and action sequence $\mathbf{s}$, generate $n$ rollouts.\footnote{We use $n=64$ for this analysis.}
    \item For each query $q$, cost-coefficients $\alpha, \beta$ find the \emph{approximately optimal} action sequence (or sequences)  $\mathbf{s}^*$, with highest average reward across rollouts.
    \item For each $\beta$ compute recall, precision, and F$_1$:
    \begin{enumerate}
        \item for each query $q$, find the optimal action sequences $s^*_1$ and $s^*_2$ for cost coefficients $(\alpha_1, \beta)$ and $(\alpha_2, \beta)$.
        \item label the query with the gold label `+C' if $s^*_2$ has a clarification question but $s^*_1$ does not, `-C', if $s^*_2$ does not have a clarification question but $s^*_1$ does, and `N' if there is no change in clarification questions.
        \item sample from the model $m$ rollouts\footnote{we use $m=256$ for this analysis.} for both $(\alpha_1, \beta)$ and $(\alpha_2, \beta)$ and estimate the \emph{predicted} action sequences $s^\textrm{model}_1$ and  $s^\textrm{model}_2$, by taking the action sequence that was the sampled the most. Similarly label each query $q$ with the predicted labels `+C', '-C', and 'N' based on the change in the predicted action sequence.
        \item Compute recall as the fraction of all examples with gold-label `+C' or '-C' for which the model predicts the correct label. Compute precision as the fraction of all examples where the model predicted `+C' or '-C' and the prediction is correct according to the gold label. Compute F$_1$ as the harmonic average.
    \end{enumerate}
    \item Average recall, precision, and F$_1$ across the values of $\beta$. 
\end{enumerate}

% Colab:
% https://colab.sandbox.google.com/drive/1ZMQ_GVXYkXpxobrLyvRi5duVZJ1S2PCL

\section{Prompts}
\label{app:prompts}

\definecolor{lightergray}{rgb}{0.9, 0.9, 0.9}

In this section we provide the prompts used in this work. 

\subsection{Filtering prompt}
\label{app:filtering_prompt}

The following prompt was used for filtering \ambigqa{} questions, where we verified that models agree with humans on whether a certain question is ambiguous or not.

\begin{tcolorbox}[
  colback=lightergray,  % Set the background color
  colframe=black,
  fontupper=\tiny\ttfamily, % Set font for the upper content part
  fontlower=\tiny\ttfamily, % Set font for the lower content part (if any)
  fonttitle=\small, % Example of styling the title differently
  title=Prompt for verifying ambiguous questions,
  breakable,
  float,
  % auto counter,
  % listtype=figure,
  label={fig:prompt_askee} 
]
You are a smart AI assistant whose goal is to answer questions. Questions can be ambiguous, that is, they might mean more than
one thing which could affect the answer. \newline

Given a question, think about all possible interpretations of this question and their answer and determine if
different interpretations lead to a different answer. If this is the case, ask a clarification question that will
allow you to answer the question. If there is just one possible interpretation or if all interpretations lead to
the same answer, just answer the question directly. \newline

Please format your thoughts and answers according to the examples below. \newline

Example 1: \newline

QUESTION: Who won the election? \newline
THOUGHT: The question doesn't say which election and different elections have
different winners so I need to ask a clarification question. \newline
CLARIFICATION: What elections are you referring to? \newline

Example 2: \newline

QUESTION: On what year did world war 2 start? \newline
THOUGHT: World war 2 is a well-defined event that started in a particular year,
1939, and so I can answer directly. \newline
ANSWER: 1939 \newline

QUESTION: <question>
\end{tcolorbox}

\newpage

\subsection{Prompts used during self-play}
\label{app:ai_assistant_prompt}

Below we provide the full prompts used during self-play. Prompts for \pacific{} are almost identical except they contain different in-context examples and include a context (document and a table).
\begin{tcolorbox}[
  colback=lightergray,  % Set the background color
  colframe=black,
  fontupper=\tiny\ttfamily, % Set font for the upper content part
  fontlower=\tiny\ttfamily, % Set font for the lower content part (if any)
  fonttitle=\small, % Example of styling the title differently
  title=Prompt for AI assistant,
  breakable,
  float,
  % auto counter,
  % listtype=figure,
  label={fig:prompt_askee} 
]
You are a helpful agent whose goal is to answer a question provided by another "asker" agent. Sometimes the question you receive might be ambiguous; if so, it can be good to ask a clarification question. You will be provided with all past clarification questions that have been asked, and their answers. As a helpful agent, your primary goal is to be accurate, while balancing asking necessary clarification questions together with providing short and succinct responses.
\newline

You can perform one of the following three actions: \newline
1. ***ANSWER** : if you are sufficiently confident that you know the meaning of the question (based on the question and all past clarification questions and their answers), you can answer it. \newline
2. **CLARIFY** : if you are unsure of what the question means (and that will affect your answer), you can ask the asker agent for more information by formulating a clarification question. \newline
3. **MULTI\_ANSWER** : if you are unsure what the question means, but you can still succinctly answer all possible interpretations of the question, you can provide a "multi-answer", that is, an answer that enumerates each interpretation and answers it. If it is not possible to enumerate a small set of interpretations, it may be better to ask a clarification question to be able to refine your answer. \newline

Let's define a few variables: \newline
`acc` is your accuracy, where 0 is not accurate at all and 100 is perfectly accurate. \newline
`num\_clar` is the number of clarification questions you had to ask. \newline
`num\_words` is the number of words in your final ANSWER assuming you chose to **ANSWER** or **MULTI\_ANSWER**.\newline

Your goal is then to maximize your `reward` which is defined as
`acc - ALPHA * num\_clar - BETA * num\_words`
where `ALPHA` and `BETA` are constants that you will be given.
In words, you want to be accurate, not ask clarification questions when they are not necessary, and you want to provide short answers. Your behavior should change depending on the particular balance of `ALPHA` vs. `BETA`, and your confidence.
Specifically, your strategy should be to estimate what action will lead to the highest reward. For example, if asking a clarification question is likely to lead to a much higher accuracy (leading to higher reward), then you should choose to ask a clarification question. Similarly, if you can already provide a multi-answer that will lead to higher accuracy and is relatively short, then you should provide a multi-answer.
\newline

**IMPORTANT NOTE**:
You should only use **MULTI\_ANSWER** if you plan to enumerate two interpretations or more and provide an answer for each interpretation. If you plan to provide just one interpretation and its answer (and optionally ask the user if they mean something else), please use **ANSWER**. \newline

Here are some examples. \newline
Example 1: \newline
ALPHA : **10.0** \newline
BETA : **1.0** \newline
QUESTION : in what year did world war ii start? \newline
ACTION : **ANSWER** : 1939 \newline

Example 2: \newline
ALPHA : **1.0** \newline
BETA : **10.0** \newline
QUESTION : in what year did the first harry potter come out? \newline
CLARIFICATION QUESTION : the book or the movie? \newline
ANSWER TO CLARIFICATION : the book. \newline
CLARIFICATION QUESTION : in what country? \newline
ANSWER TO CLARIFICATION : the uk. \newline
ACTION : **ANSWER** : 1997 \newline

Example 3: \newline
ALPHA : **10.0** \newline
BETA : **0.1** \newline
QUESTION : in what year did the first harry potter come out? \newline
CLARIFICATION QUESTION : the book or the movie? \newline
ANSWER TO CLARIFICATION : the book. \newline
ACTION : **MULTI\_ANSWER** : The book came out in the uk in 1997 and in the us in 1998. \newline

Example 4: \newline
ALPHA : **10.0** \newline
BETA : **0.1** \newline
QUESTION : in what year did the first harry potter come out in the US? \newline
ACTION : **MULTI\_ANSWER** : The first harry potter book came out in the us in 1998,the first harry potter movie in 2001. \newline

Example 5: \newline
ALPHA : **0.1** \newline
BETA : **1.0** \newline
QUESTION : in what year did the first harry potter come out? \newline
CLARIFICATION QUESTION : the book or the movie? \newline
ANSWER TO CLARIFICATION : the book. \newline
ACTION : **CLARIFY** : in the uk or the us? \newline

ALPHA : \{ALPHA\_VAR\} \newline
BETA : \{BETA\_VAR\} \newline
QUESTION : \{AMBIG\_QUERY\_VAR\}
\{OPTIONAL\_CLARIFICATION\_QUESTIONS\_VAR\}
\end{tcolorbox}

\newpage

\begin{tcolorbox}[
  colback=lightergray,  % Set the background color
  colframe=black,
  fontupper=\tiny\ttfamily, % Set font for the upper content part
  fontlower=\tiny\ttfamily, % Set font for the lower content part (if any)
  fonttitle=\small, % Example of styling the title differently
  title=Prompt for AI assistant with chain-of-thought,
  breakable,
  float,
  % auto counter,
  % listtype=figure,
  label={fig:prompt_askee_cot} 
  ]
You are a helpful agent whose goal is to answer a question provided by another "asker" agent. Sometimes the question you receive might be ambiguous; if so, it can be good to ask a clarification question. You will be provided with all past clarification questions that have been asked, and their answers. As a helpful agent, your primary goal is to be accurate, while balancing asking necessary clarification questions together with providing short and succinct responses.
 \newline

You can perform one of the following three actions: \newline
1. ***ANSWER** : if you are sufficiently confident that you know the meaning of the question (based on the question and all past clarification questions and their answers), you can answer it. \newline
2. **CLARIFY** : if you are unsure of what the question means (and that will affect your answer), you can ask the asker agent for more information by formulating a clarification question. \newline
3. **MULTI\_ANSWER** : if you are unsure what the question means, but you can still succinctly answer all possible interpretations of the question, you can provide a "multi-answer", that is, an answer that enumerates each interpretation and answers it. If it is not possible to enumerate a small set of interpretations, it may be better to ask a clarification question to be able to refine your answer.
You will be provided with a list of ALLOWED\_ACTIONS, since not all actions are available in all situations. \newline

Let's define a few variables: \newline
`acc` is your accuracy, where 0 is not accurate at all and 100 is perfectly accurate. \newline
`num\_clar` is the number of clarification questions you had to ask. \newline
`num\_words` is the number of words in your final ANSWER assuming you chose to **ANSWER** or **MULTI\_ANSWER**. \newline

Your goal is then to maximize your `reward` which is defined as
`acc - ALPHA * num\_clar - BETA * num\_words`
where `ALPHA` and `BETA` are constants that you will be given.
In words, you want to be accurate, not ask clarification questions when they are not necessary, and you want to provide short answers. Your behavior should change depending on the particular balance of `ALPHA` vs. `BETA`, your confidence, and the set of ALLOWED\_ACTIONS.
Specifically, your strategy should be to estimate which of the ALLOWED\_ACTIONS will lead to the highest reward. For example, if asking a clarification question is likely to lead to a much higher accuracy (that will lead to higher reward), then you should choose to ask a clarification question. Similarly, if you can already provide a multi-answer that will lead to higher accuracy and is relatively short, then you should choose to provide a multi-answer. To estimate this, you should output a THOUGHT where you try to explicitly estimate the reward for different actions. After your THOUGHT is done, you should output the ACTION that you will take, as shown in the examples. \newline

Here are some examples. \newline
Example 1: \newline
ALPHA : **10.0** \newline
BETA : **1.0** \newline
ALLOWED\_ACTIONS : \{`CLARIFY', `MULTI\_ANSWER', `ANSWER'\} \newline
QUESTION : in what year did world war ii start? \newline
THOUGHT : I am confident that the answer is 1939, and thus I can answer the question correctly with one word. My reward will be 100.0 - 10.0 * 0 - 1.0 * 1 = 99.0. This is the highest reward I can get, and so I will choose to answer directly. \newline
ACTION : **ANSWER** : 1939 \newline

Example 2: \newline
ALPHA : **1.0** \newline
BETA : **10.0** \newline
ALLOWED\_ACTIONS : \{`MULTI\_ANSWER', `ANSWER'\} \newline
QUESTION : in what year did the first harry potter come out? \newline
CLARIFICATION QUESTION : the book or the movie? \newline
ANSWER TO CLARIFICATION : the book. \newline
CLARIFICATION QUESTION : in what country? \newline
ANSWER TO CLARIFICATION : the uk. \newline
THOUGHT : Given the previous clarification questions and answers, I am confident that the answer is 1997, and thus I can answer the question correctly with one word. Moreover, I cannot ask additional clarification questions because CLARIFY is not in the list of ALLOWED\_ACTIONS. I already asked two clarification questions in previous turns, so my reward will be 100.0 - 1.0 * 2 - 10.0 * 1 = 88.0. This is the highest reward I can get at this point, so I will choose to answer. \newline
ACTION : **ANSWER** : 1997 \newline

Example 3: \newline
ALPHA : **10.0** \newline
BETA : **0.1** \newline
ALLOWED\_ACTIONS : \{`MULTI\_ANSWER', `ANSWER'\} \newline
QUESTION : in what year did the first harry potter come out? \newline
CLARIFICATION QUESTION : the book or the movie? \newline
ANSWER TO CLARIFICATION : the book. \newline
THOUGHT : I know that the question is about the book, however, I am not sure if the question is about the publication date of the book in the uk or in the us. If I guess one of the two and answer directly, I can answer in one word. My expected accuracy will be 50.0. Since I already asked one clarification question, my reward will be 50.0 - 10.0 * 1 - 0.1 * 1 = 39.9. I cannot ask a clarification question because it is not in the list of ALLOWED\_ACTIONS. Finally, if I provide a multi-answer that contains the dates for both the uk and the us, my accuracy will be 100.0, but my answer will be long; I expect around 10 words. Since I already asked one clarification question and my response will be ~10 words, my reward will be 100.0 - 10.0 * 1 - 0.1 * 10 = 89.0. This is the best expected score I can get, because the cost of clarification questions is high relative to the cost of words in my final answer. \newline
ACTION : **MULTI\_ANSWER** : The book came out in the uk in 1997 and in the us in 1998. \newline

Example 4: \newline
ALPHA : **10.0** \newline
BETA : **0.1** \newline
ALLOWED\_ACTIONS : \{`CLARIFY', `MULTI\_ANSWER', `ANSWER'\} \newline
QUESTION : in what year did the first harry potter come out in the US? \newline
THOUGHT : I am not sure if the question is about the movie or the book? If I guess one of the two and answer directly, I can answer in one word. My expected accuracy will be 50.0. Since I have not asked any clarification questions yet, my reward will be 50.0 - 10.0 * 0 - 0.1 * 1 = 49.9. If I ask a clarification question, I will be sure of the correct answer, I can answer in one word, and my accuracy will be 100.0. Since I will have asked one clarification question in total, my reward will be 100.0 - 10.0 * 1 - 0.1 * 1 = 89.9. Finally, if I provide a multi-answer that covers both the movie and the book, my accuracy will be 100.0, but my answer will be long; I expect around 15 words. Since I have not asked any clarification questions yet, my reward will be 100.0 - 10.0 * 0 - 0.1 * 15 = 98.5. This is the best expected score I can get, because the cost of clarification questions is high relative to the cost of words in my final answer.
ACTION : **MULTI\_ANSWER** : The first harry potter book came out in the us in 1998,the first harry potter movie in 2001. \newline

Example 5: \newline
ALPHA : **0.1** \newline
BETA : **1.0** \newline
ALLOWED\_ACTIONS : \{`CLARIFY', `MULTI\_ANSWER', `ANSWER'\} \newline
QUESTION : in what year did the first harry potter come out? \newline
CLARIFICATION QUESTION : the book or the movie? \newline
ANSWER TO CLARIFICATION : the book. \newline
THOUGHT : Given the previous clarification questions and answers, I am still not sure if the question is about publication date in the uk or the us. If I guess one of the two and answer directly, I can answer in one word. My expected accuracy will be 50.0, and my reward will be 50.0 - 0.1 * 1 - 1.0 * 1 = 48.9. If I ask a clarification question my accuracy will be 100.0, and my reward will be 100.0 - 0.1 * 2 - 1.0 * 1 = 98.8. If I provide a multi-answer, my accuracy will be 100.0, but my answer will be long, around 15 words, so my reward will be 100.0 - 0.1 * 1 - 1.0 * 15 = 84.9. So because the cost of clarification questions is low relative to the cost of words in my final answer, I will choose to ask a clarification question. \newline
ACTION : **CLARIFY** : in the uk or the us? \newline

ALPHA : \{ALPHA\_VAR\} \newline
BETA : \{BETA\_VAR\} \newline
ALLOWED\_ACTIONS : \{ALLOWED\_ACTIONS\_VAR\} \newline
QUESTION : \{AMBIG\_QUERY\_VAR\}\ {OPTIONAL\_CLARIFICATION\_QUESTIONS\_VAR\}}
\end{tcolorbox}

 \newpage
\begin{tcolorbox}[
  colback=lightergray,  % Set the background color
  colframe=black,
  fontupper=\tiny\ttfamily, % Set font for the upper content part
  fontlower=\tiny\ttfamily, % Set font for the lower content part (if any)
  fonttitle=\small, % Example of styling the title differently
  title=Prompt for user simulator answering a clarification question,
  breakable,
  float,
  % auto counter,
  % listtype=figure,
  label={fig:filtering} 
]
You are a helpful agent whose goal is to clarify a potentially ambiguous question. You have access to a question in two forms. One is a potentially ambiguous form and the other is an unambiguous form of that question. \newline

You will also be given a clarification question issued by a different agent that only has access to the ambiguous question and is trying to clarify it. Your only job is to answer the clarification question to the best of your ability. It is possible that the ambiguous and unambiguous questions are identical. \newline

You are not allowed to directly reveal information about the unambiguous question. You are only allowed to answer the clarification question based on  the unambiguous version of the question. If the clarification question is not answerable from the unambiguous question, you should answer that you do not know the answer. Again, **NEVER** just repeat the unambiguous question as an answer. In addition, **NEVER** clarify the question using information that is not inferred from the unambiguous question. You will get a prize if you manage to complete the task successfully. Always respond in a single line with the format \newline
'ACTION : **ANSWER\_CLARIFICATION** : <the answer to the clarification question>' \newline

Example 1, where the unambiguous question contains information about the answer: \newline
Potentially ambiguous question: When did the first harry potter come out? \newline
Unambiguous question: When did the first harry potter book come out in the uk? \newline
Clarification question: Did you mean the book or the movie? \newline
ACTION : **ANSWER\_CLARIFICATION** : the book. \newline

Example 2, where the unambiguous question contains information about the answer: \newline
Potentially ambiguous question: When did the first harry potter come out? \newline
Unambiguous question: When did the first harry potter book come out in the uk? \newline
Clarification question: In which country? \newline
ACTION : **ANSWER\_CLARIFICATION** : the uk. \newline

Example 3, where the unambiguous question does not contain information about the answer: \newline
Potentially ambiguous question: When did the first harry potter come out? \newline
Unambiguous question: When did the first harry potter book come out in the uk? \newline
Clarification question: did you mean hard cover or soft cover? \newline
ACTION : **ANSWER\_CLARIFICATION** : I do not know. \newline

Now for the real data: \newline
Potentially ambiguous question: {AMBIG\_QUERY\_VAR} \newline
Unambiguous question: {UNAMBIG\_QUERY\_VAR} \newline
Clarification question: {CLARIF\_QUERY\_VAR} 
\end{tcolorbox}

\newpage

\begin{tcolorbox}[
  colback=lightergray,  % Set the background color
  colframe=black,
  fontupper=\tiny\ttfamily, % Set font for the upper content part
  fontlower=\tiny\ttfamily, % Set font for the lower content part (if any)
  fonttitle=\small, % Example of styling the title differently
  title=Prompt for user simulator returning a final answer,
  breakable,
  float,
  % auto counter,
  % listtype=figure,
  label={fig:filtering} 
]
You are a helpful agent whose goal is to take a question and an answer to the question that is provided to you by another agent, and use it to phrase a final answer to the question. \newline

You will get two versions of the question, one is a potentially ambiguous question, and the other is an unambiguous version. You will also get the answer provided by the other agent. The other agent does not have direct access to the unambiguous question. Your goal is to answer the question based on the answer provided by the other agent and your privileged knowledge of the unambiguous question. DO NOT answer the question based on your own knowledge. Only use the answer given by the other agent, and your knowledge about the unambiguous question. \newline

You will use the action **ANSWER** to pass along the answer. \newline

Importantly, the other agent can give an answer in one of two forms: \newline
1. Just give a single answer to the question, in which case you just need to pass it along as is. \newline
2. The other agent might give you a `multi-answer` where it specifies possible interpretations to the question and the answer to each interpretations. Since you have access to the unambiguous version you can select the right interpretation and provide an answer based on that. \newline

Always respond in a single line with the format
'ACTION : **ANSWER** : <the answer to the question>' \newline

Example 1, where the answer given by the other agent is wrong and is passed as is: \newline
Potentially ambiguous question: in what year did world war i start? \newline
Unambiguous question: in what year did world war i start?
FRIEND's ANSWER: 1939 \newline
ACTION : **ANSWER** : 1939 \newline

Example 2, where the answer given by the other agent is correct and is passed as is: \newline
Potentially ambiguous question: in what year did world war ii start? \newline
Unambiguous question: in what year did world war ii start? \newline
FRIEND's ANSWER: 1939 \newline
ACTION : **ANSWER** : 1939 \newline

Example 3, where the other agent provides a multi-answer: \newline
Potentially ambiguous question: in what year did the first harry potter come out? \newline
Unambiguous question: in what year did the first harry potter book come out in the uk? \newline
FRIEND's ANSWER: If you mean when the book harry potter came out in the uk, 
then the answer is 1997, if you mean when did the first harry potter
book come out in the us, the answer is 1998. If you mean when did the first 
harry potter movie come out, then the answer is 2001. \newline
ACTION : **ANSWER** : 1997 \newline

Example 4, where the other agent provides an answer: \newline
Potentially ambiguous question: in what year did the first harry potter come out? \newline
Unambiguous question: in what year did the first harry potter book come out in the uk? \newline
FRIEND's ANSWER: I think you mean when the movie came out, and that was in 2001. \newline
ACTION : **ANSWER** : 2001. \newline

Now for the real data: \newline
Potentially ambiguous question: {AMBIG\_QUERY\_VAR} \newline
Unambiguous question: {UNAMBIG\_QUERY\_VAR} 
\end{tcolorbox}

\end{document}